\documentclass[a4paper,fleqn]{cas-sc}

\usepackage[authoryear,longnamesfirst]{natbib}
\usepackage{algorithm}
\usepackage{algpseudocode}
\usepackage{booktabs}
\usepackage{multirow}
\usepackage{amsmath}
\usepackage{amssymb}
\usepackage{bm}
\usepackage{enumitem}
\usepackage{verbatim}
\usepackage{algorithm}
\usepackage{algpseudocode}
\usepackage{float}
\def\tsc#1{\csdef{#1}{\textsc{\lowercase{#1}}\xspace}}
\tsc{WGM}
\tsc{QE}

\begin{document}
\let\WriteBookmarks\relax

\shorttitle{Neuro‑Symbolic LEED Compliance}    

\shortauthors{Aritro De; Juliana Felkner}  

\title [mode = title]{Neuro‑Symbolic AI for LEED compliance: Document‑Centric Benchmarking, Deterministic Numeric Checking, and When Multimodal Hurts}  

\tnotemark[1] 

\tnotetext[1]{} 

%

\author[1]{Aritro De}

\cormark[1]

\fnmark[1]

\ead{aritro@utexas.edu}



\affiliation[1]{organization={The University of Texas at Austin},
            addressline={}, 
            city={Austin},
            citysep={}, 
            postcode={78712}, 
            state={Texas},
            country={United States of America}}

\author[1]{Juliana Felkner}

\fnmark[2]








\begin{abstract}
LEED v4.1 BD+C certification remains a document-intensive process that requires reviewers to read hundreds of pages of project evidence and apply credit-specific threshold logic by hand. This paper investigates whether small, locally deployed language models can perform meaningful screening of LEED documentation and how deterministic symbolic components should share that work. A neuro-symbolic pipeline is introduced that aligns project PDFs to LEED credit sections, retrieves evidence with credit-aware keyword signatures, verifies compliance with a locally hosted 4-billion-parameter language model, and applies a LEED-specific numeric checker to quantitative thresholds.

Experiments on four university buildings (484 PDFs, 153 credit-level decisions) show that a 4-billion-parameter model (gemma3:4b) is the strongest text-only core verifier, achieving 67.3\% accuracy and outperforming a larger 8-billion-parameter model (llama3.1:8b) in this task. The deterministic numeric checker corrects arithmetic errors on key quantitative credits, moving EA-p2 from 50.0\% to 100.0\% accuracy and improving several other credits when required values are reliably extracted. At the same time, the full neuro-symbolic configuration achieves 61.6\% overall accuracy, trailing the best text-only baseline due to extraction failures and conservative behavior on qualitative categories.

Systematic ablations show that adding low-resolution drawing images (150–300 dpi) consistently reduces accuracy, and that prompt effectiveness depends on the building’s ground-truth PASS rate: rubric prompts perform best on documentation-rich projects, while chain-of-thought prompts perform best on documentation-lean projects. Within the specific scope of LEED v4.1 BD+C compliance verification over raw project documentation, this pipeline and its baselines provide an initial reproducible reference point for both accuracy and failure modes. The results support a pragmatic design: use a strong local text-only model as the default verifier, add deterministic numeric checking selectively for high-consequence quantitative credits, tune prompts to expected documentation strength, and treat extraction as the primary engineering frontier.
\end{abstract}


\begin{highlights}
\item Presents the first neuro-symbolic pipeline for LEED v4.1 BD+C compliance over real project documentation, with credit-section-aware chunking and agentic retrieval.
\item Implements a LEED-specific deterministic numeric checker that corrects arithmetic errors on high-consequence quantitative credits, including EA-p2 moving from 50\% to 100\% accuracy.
\item Benchmarks locally deployed small language models on four UT Austin buildings, showing a 4B model outperforming an 8B model and documenting prompt sensitivity to class imbalance.
\item Provides systematic multimodal and extraction ablations, demonstrating that low-resolution drawing images and imperfect numeric extraction can reduce accuracy and identifying where symbolic and neural components fail.
\end{highlights}


\maketitle

\section{Introduction}
\label{section:intro}

Buildings certified under the Leadership in Energy and Environmental Design (LEED) System must undergo a labor-intensive document review process that relies on extensive data handling, simulation outputs, and documentation review \citep{lee_integrated_2025}. Reviewers must then align heterogeneous project submissions with the requirements of each attempted credit, making certification as much an evidence-matching task as a design evaluation task \citep{feijao_comparative_2024}. Research on automated compliance checking has been active for decades in construction, beginning with rule-based systems that translated regulations into machine-readable logic.

Automated compliance checking has evolved from rigid rule-based engines to hybrid systems that combine symbolic logic, BIM, and NLP. Early work translated regulatory clauses into computable rules, but these methods required large amounts of manual formalization and were difficult to adapt across jurisdictions. More recent NLP work has aimed to extract semantic elements and relations from regulatory text so that rules can be generated more flexibly \citep{zhang_semantic_2016,hettiarachchi_code-accord_2025,lee_automated_2026,mirhosseini_systematic_2026}.

A 2026 systematic review of methods for interpreting building code regulations in automated compliance systems found that the field is moving toward mixed architectures that combine language understanding with structured checking pipelines \citep{mirhosseini_systematic_2026}. Related work has shown that deep learning and grammar-based parsing can outperform older handcrafted approaches in extracting regulatory logic from complex sentences. At the same time, automated compliance checking across the building lifecycle remains limited by document complexity, data heterogeneity, and the challenge of turning natural language into reliable computable evidence \citep{hettiarachchi_code-accord_2025,lee_automated_2026}.

Recent studies have begun using LLMs directly in compliance workflows. BIM-linked systems have used GPT, Claude, Gemini, and Llama to interpret code requirements and generate semi-automated checking routines inside design environments \citep{madireddy_large_2025}. The U.S. Department of Energy has also explored AI-assisted energy code compliance verification, pairing code retrieval with a question-answering interface to support more transparent review \citep{wan_exploring_2025}.

Emerging research on LEED automation suggests that AI could reduce the time and effort tied to this process. One recent integrated platform for LEED certification automation reports significant reductions in documentation effort, but it focuses more on automation architecture than on rigorous cross-building, credit-level benchmarking \citep{lee_integrated_2025}. That leaves open a core research question: how accurate can a small, local model actually be when asked to verify credit evidence across real project submittals?

Retrieval-augmented generation is now a standard way to answer questions over domain-specific corpora \citep{oche_systematic_2025}. But domain-specific corpora with stable and technical vocabularies often reward high-precision keyword retrieval instead \citep{jadon_enhancing_2025}. Self-consistency is also relevant. Prior work shows that generating multiple reasoning paths and selecting the majority answer can reduce stochastic error and improve reliability \citep{taubenfeld_confidence_2025, huang_mirror-consistency_2024}. Calibration research likewise shows that larger models are not always better behaved \citep{michael_confidence_2026}. Overconfidence and hallucination remain serious issues, especially when models are rewarded for always producing an answer instead of abstaining when evidence is weak \citep{kalai_why_2025,geng_survey_2024}. Those concerns map directly onto LEED review, where the correct answer is often not PASS or FAIL, but simply that the evidence is insufficient.

This work makes three specific contributions. First, it proposes a neuro‑symbolic framework for LEED v4.1 BD+C compliance verification that operates on real project documentation rather than on curated BIM or IFC models. The pipeline aligns heterogeneous PDFs with credit sections, retrieves evidence using credit‑aware signatures, and separates narrative reasoning from deterministic threshold checking. Second, it defines and implements a LEED‑specific numeric checker that encodes quantitative thresholds and point tables and shows that separating arithmetic from language improves correctness on high‑consequence credits such as EA‑p2. Third, it develops a multi‑building benchmark for LEED document‑level automated compliance checking, reporting per‑category accuracy, multimodal ablation, prompt‑sensitivity analysis, and an error taxonomy that captures both positive and negative interactions between neural and symbolic components. Taken together, these contributions move LEED compliance from a black‑box application of general language models toward a structured, inspectable baseline that can support cumulative methodological development.

Within the specific scope of LEED v4.1 BD+C compliance verification over raw project documentation, this pipeline and its baselines offer, to the best of available evidence, the most complete published treatment to date. In the absence of prior neuro‑symbolic systems or open benchmarks for LEED document‑centric automated compliance checking, the study establishes an initial reproducible reference point, including quantitative accuracy and a qualitative analysis of failure modes.

\section{Methodology}
\label{sec:methodology}

\subsection{Research Design and Pipeline Overview}
\label{sec:overview}

This study develops and evaluates a neuro-symbolic system for verifying whether project documentation supports compliance with LEED v4.1 Building Design and Construction (BD+C) requirements. The system is designed for a difficult but common condition in green-building certification: evidence is distributed across a heterogeneous set of unstructured documents, including narratives, schedules, energy reports, commissioning plans, material submittals, construction waste reports, and drawings. The system does not assume that project evidence has been converted into a complete BIM model or a normalized database. It begins with the documents that project teams actually submit.

The central design principle is separation of concerns. The language model is used for tasks that require contextual interpretation: locating potentially relevant text, extracting evidence, identifying numerical values, and explaining a decision. A deterministic rule engine is used for tasks that require exact arithmetic or threshold comparison. This separation matters because an apparently plausible language-model explanation does not guarantee a correct numerical conclusion. In the proposed pipeline, a model may identify an energy-model result, but it does not decide whether the reported value clears the relevant LEED threshold. The numeric checker does that work.

\begin{figure*}[H]
\centering
\includegraphics[width=0.95\textwidth]{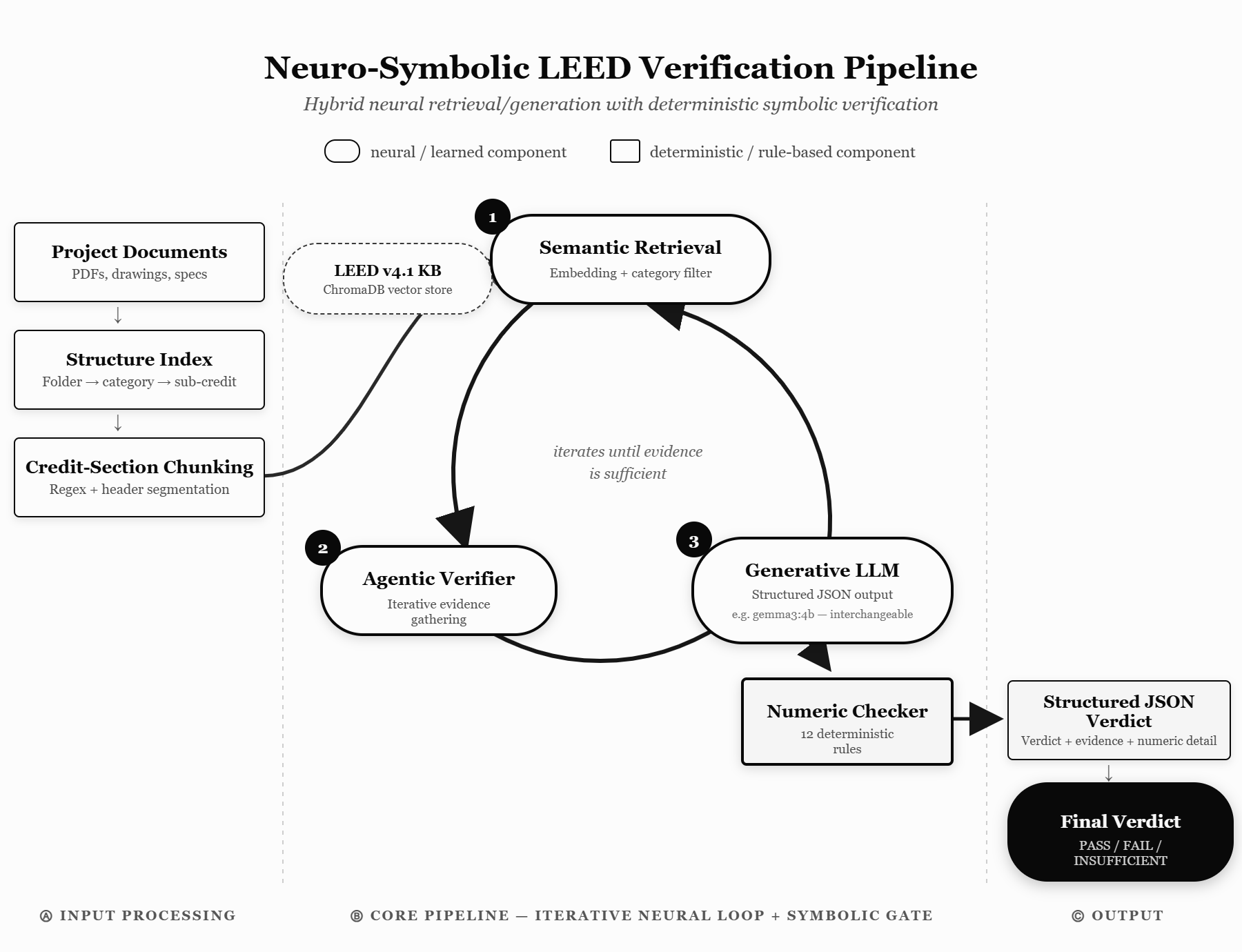}
\caption{Neuro-symbolic LEED verification pipeline. The pipeline transforms heterogeneous project documentation into credit-level verdicts. The language model performs retrieval-guided evidence interpretation and value extraction. The symbolic checker applies deterministic credit-specific threshold logic. The checker overrides the language-model result only when it returns a determinate PASS or FAIL verdict.}
\label{fig:architecture}
\end{figure*}

Figure~\ref{fig:architecture} summarizes the seven-stage pipeline. The process begins by constructing a searchable representation of the LEED submission directory. It then identifies credit-specific document sections, retrieves candidate evidence, obtains a structured language-model assessment, applies calibrated promotion and deterministic numeric checks, and returns a traceable final result. Each result contains a verdict, confidence score, evidence quotations, missing evidence elements, numerical values, and a record of whether deterministic logic changed the original model decision.

Let the input corpus for building $b$ be represented as

\begin{equation}
\mathcal{D}^{(b)} = \{d_1, d_2, \dots, d_N\},
\label{eq:corpus_definition}
\end{equation}

where each $d_i$ is a PDF document associated with the building. For every LEED credit $c \in \mathcal{L}$, where $\mathcal{L}$ is the evaluated credit set, the objective is to produce a final verdict

\begin{equation}
\hat{y}_{b,c} \in \{\textsc{pass}, \textsc{fail}, \textsc{insufficient\_data}\}.
\label{eq:verdict_space}
\end{equation}

The system produces this verdict from retrieved evidence $E_{b,c}$, a language-model assessment $z_{b,c}$, a calibrated promotion function $P(\cdot)$, and, where applicable, a deterministic rule function $R_c(\cdot)$:

\begin{equation}
\hat{y}_{b,c} =
R_c\left(
P\left(z_{b,c}, \tau_b\right),
E_{b,c}
\right).
\label{eq:final_decision}
\end{equation}

The promotion threshold $\tau_b$ may be fixed globally or calibrated using held-out development data, as described in Section~\ref{sec:promotion}. The rule function $R_c(\cdot)$ only replaces a language-model verdict when it can establish a determinate \textsc{pass} or \textsc{fail} result. If the rule engine cannot recover the required values or establish a valid comparison, the prior language-model verdict is preserved.

\subsection{Document Ingestion and Structure Indexing}
\label{sec:indexing}

LEED project submissions are not a flat collection of files. They commonly preserve a partial organizational structure created by the project team, consultant, or LEED administrator. This structure often includes top-level category folders such as Energy and Atmosphere (EA), Water Efficiency (WE), Sustainable Sites (SS), Materials and Resources (MR), Indoor Environmental Quality (EQ), Innovation (IN), and Regional Priority (RP). Within these folders, project teams frequently create subdirectories for individual prerequisites or credits. The resulting hierarchy provides useful but imperfect metadata. It is useful because folder location can narrow the retrieval space. It is imperfect because teams use inconsistent abbreviations, legacy names, and non-standard directory labels.

The \texttt{StructureIndex} converts this heterogeneous directory structure into a normalized hierarchy:

\begin{verbatim}
Building
  -> LEED Category
      -> Sub-credit Folder
          -> PDF Document
              -> Page Text and Metadata
\end{verbatim}

Each PDF is processed using PyMuPDF (\texttt{fitz}). The ingestion stage extracts page-level text, page count, source filename, absolute and relative path, document title when available, and file-level metadata. Extracted text is cached locally so that subsequent retrieval operations do not require repeated PDF parsing. Caching is important because a single building may contain more than one hundred documents and repeated extraction would otherwise dominate runtime.

For each document $d_i$, the index stores a structured record

\begin{equation}
I(d_i) =
\left[
t_i,\,
p_i,\,
m_i,\,
g_i,\,
s_i
\right],
\label{eq:document_index}
\end{equation}

where $t_i$ is extracted full text, $p_i$ is page-level text and page count, $m_i$ is file metadata, $g_i$ is the normalized LEED category inferred from the directory path, and $s_i$ is the candidate sub-credit label inferred from the source directory or filename.

The category normalization step maps directory names to canonical LEED category codes. The mapping is intentionally conservative. Direct matches such as \texttt{EA}, \texttt{WE}, \texttt{SS}, and \texttt{MR} are retained. Legacy or project-specific labels are translated only when the correspondence is unambiguous. For example, \texttt{ID} is mapped to Innovation (\texttt{IN}), \texttt{IQ} is mapped to Indoor Environmental Quality (\texttt{EQ}), and \texttt{PI} is mapped to Integrative Process (\texttt{IP}). Table~\ref{tab:dir_category} lists the complete mapping used in the study.

\begin{table}[htbp]
\centering
\small
\caption{Directory-to-category normalization used by the structure index.}
\label{tab:dir_category}
\begin{tabular}{lll}
\toprule
\textbf{Directory Label} & \textbf{Canonical Category} & \textbf{Interpretation} \\
\midrule
EA, WE, SS, MR & EA, WE, SS, MR & Direct category match \\
LT, IN, IP, RP & LT, IN, IP, RP & Direct category match \\
ID & IN & Innovation and Design abbreviation \\
IQ & EQ & Indoor Environmental Quality abbreviation \\
PI & IP & Process and Integrative Process abbreviation \\
\bottomrule
\end{tabular}
\end{table}

The normalized category is used as a retrieval prior. For a target credit $c$ belonging to category $g(c)$, the first-pass search universe is

\begin{equation}
\mathcal{D}_{g(c)} =
\left\{
d_i \in \mathcal{D}^{(b)} : g_i = g(c)
\right\}.
\label{eq:category_filter}
\end{equation}

Category pinning reduces the number of candidate documents substantially while retaining the most likely evidence source. It also reduces the chance that generic sustainability language from unrelated documents is treated as credit-specific evidence. However, category filtering is not treated as an absolute constraint. If the initial search fails to recover sufficient evidence, the agent may expand retrieval to neighboring categories or the full project corpus. This fallback is necessary because LEED evidence is often cross-cutting. For example, a commissioning document may support both Energy and Atmosphere and Indoor Environmental Quality credits.

\subsection{Credit-Section-Aware Chunking}
\label{sec:chunking}

Retrieval quality depends not only on the embedding model or search method but also on the unit of text that is retrieved. Fixed-length chunking is convenient, but it is poorly matched to certification documentation. It may separate a requirement from the calculation that proves it, split a table explanation across chunks, or combine unrelated credit discussions in one retrieval unit. These failures are especially harmful when the downstream task is binary or ternary compliance verification rather than general summarization.

The proposed method uses credit-section-aware chunking. It attempts to identify document segments that correspond to one LEED credit or prerequisite. The method searches the extracted text for three classes of headings and aliases:

\begin{enumerate}
    \item \textbf{LEED v4.1 identifiers}, such as ``EA Credit 1,'' ``WE Prerequisite 2,'' and ``MR Credit 5.''
    \item \textbf{LEED 2009 identifiers}, such as ``EAc1,'' ``WEp2,'' and ``MRc5,'' which remain common in documentation from projects that transitioned between rating-system versions.
    \item \textbf{Credit-name aliases}, such as ``Fundamental Commissioning,'' ``Optimize Energy Performance,'' ``Indoor Water Use Reduction,'' and ``Construction and Demolition Waste Management.''
\end{enumerate}

The alias dictionary contains more than 80 normalized patterns. It maps an observed document heading or phrase to a canonical credit identifier. For example, \texttt{fundamental commissioning} maps to \texttt{EA-p1}, while \texttt{optimize energy performance} maps to \texttt{EA-c1}. This mapping preserves continuity across rating-system terminology, project-specific naming, and abbreviated folder labels.

Let $\mathcal{P} = \{(p_j, c_j)\}_{j=1}^{J}$ denote the set of regular-expression patterns and their associated canonical credit identifiers. For each line $\ell_r$ in document $d_i$, the section-label function is

\begin{equation}
h(\ell_r) =
\begin{cases}
c_j, & \text{if } \exists (p_j,c_j) \in \mathcal{P} \text{ such that } p_j \text{ matches } \ell_r, \\
\varnothing, & \text{otherwise}.
\end{cases}
\label{eq:header_detection}
\end{equation}

When $h(\ell_r) \neq \varnothing$, the system closes the prior section and begins a new section associated with the detected credit. Documents without recognized headers are retained as document-level chunks with category and source-file metadata. This fallback prevents loss of relevant evidence from narrative reports that do not follow LEED naming conventions.

A section longer than 3,000 characters is split at the nearest paragraph boundary rather than at a fixed character boundary. The split preserves the same canonical credit identifier and records whether the chunk is a complete section or a continuation. Each final chunk $q_k$ is represented as

\begin{equation}
q_k =
\left[
x_k,\,
c_k,\,
g_k,\,
f_k,\,
\pi_k,\,
\Lambda_k,\,
\omega_k
\right],
\label{eq:chunk_representation}
\end{equation}

where $x_k$ is chunk text, $c_k$ is an explicit credit identifier when detected, $g_k$ is normalized category, $f_k$ is source filename, $\pi_k$ is the page range, $\Lambda_k$ is the set of likely categories or credits, and $\omega_k$ indicates whether the chunk represents a complete section or an overflow continuation.

This representation makes the provenance of every retrieved passage available to the verifier and to the end user. A verdict is therefore not simply associated with a document. It can be traced to a specific chunk, source file, and page range.

\begin{algorithm}[htbp]
\caption{Credit-Section-Aware Document Chunking}
\label{alg:chunking}
\begin{algorithmic}[1]
\Require Extracted document text $T$, pattern map $\mathcal{P}$, maximum length $L_{\max}$
\Ensure Credit-aware chunk set $\mathcal{Q}$

\State Split $T$ into ordered lines $\{\ell_1,\ell_2,\dots,\ell_R\}$
\State Initialize $\mathcal{Q} \gets \emptyset$, $B \gets \emptyset$, $c_{\mathrm{active}} \gets \varnothing$

\For{$r = 1$ to $R$}
    \State $c_{\mathrm{new}} \gets h(\ell_r)$ \Comment{via Equation~\ref{eq:header_detection}}
    \If{$c_{\mathrm{new}} \neq \varnothing$}
        \If{$B \neq \emptyset$}
            \State Create chunk from $B$ with metadata $c_{\mathrm{active}}$
            \State Add chunk to $\mathcal{Q}$
        \EndIf
        \State $B \gets \{\ell_r\}$ and $c_{\mathrm{active}} \gets c_{\mathrm{new}}$
    \Else
        \State Append $\ell_r$ to $B$
    \EndIf
\EndFor

\If{$B \neq \emptyset$}
    \State Create final chunk from $B$ and add it to $\mathcal{Q}$
\EndIf

\For{each chunk $q \in \mathcal{Q}$}
    \If{$|x_q| > L_{\max}$}
        \State Split $q$ at nearest paragraph boundaries while preserving metadata
    \EndIf
\EndFor

\State \Return $\mathcal{Q}$
\end{algorithmic}
\end{algorithm}

\subsection{Three-Tier Evidence Retrieval}
\label{sec:retrieval}

The retrieval system is designed around a practical observation: LEED documentation contains highly specialized, stable vocabulary. Terms such as \textit{sDA}, \textit{EUI}, \textit{SRI}, \textit{gpf}, \textit{gpm}, \textit{OPR}, and \textit{Basis of Design} often have stronger evidentiary value than broad semantic similarity. A passage may be semantically related to water efficiency but still fail to state fixture flow rates, baseline calculations, or percentage reduction. For verification, topical relevance is not enough. The system must retrieve evidence that can support a decision.

The retrieval process has three tiers. The first tier applies credit-specific lexical retrieval to cached full text. Each credit $c$ is associated with a keyword signature $W_c$. For example, EA-p1 uses terms such as \textit{commissioning}, \textit{Cx}, \textit{ASHRAE Guideline}, \textit{Owner's Project Requirements}, and \textit{Basis of Design}. WE-p2 uses \textit{fixture}, \textit{gpf}, \textit{gpm}, \textit{flow rate}, and \textit{indoor water}. MR-c5 uses \textit{construction waste}, \textit{diversion}, \textit{recycle}, and \textit{landfill}. The second tier ranks credit-aware document chunks using metadata and lexical evidence. The third tier uses semantic retrieval as a fallback when exact vocabulary is weak, inconsistent, or absent.

For a target credit $c$ and candidate chunk $q_k$, the lexical relevance score is

\begin{equation}
S_{\mathrm{lex}}(c,q_k)
=
\sum_{w \in W_c}
2 \cdot |w|_{\mathrm{tok}}
\cdot \mathbb{1}[w \in x_k]
+
5 \cdot \mathbb{1}[c \in \Lambda_k],
\label{eq:lexical_score}
\end{equation}

where $|w|_{\mathrm{tok}}$ is the number of tokens in keyword phrase $w$, $x_k$ is the chunk text, and $\Lambda_k$ is the set of likely credits associated with the chunk. Multi-word terms receive larger weights because a phrase such as ``Basis of Design'' is more diagnostic than a single generic word such as ``design.'' The metadata bonus rewards chunks already associated with the target credit through section detection, directory location, or filename evidence.

For semantic retrieval, each chunk is embedded using \texttt{all-MiniLM-L6-v2}, which maps text to a 384-dimensional dense vector representation. Let $\mathbf{e}(x_k) \in \mathbb{R}^{384}$ denote a chunk embedding and $\mathbf{e}(r_c) \in \mathbb{R}^{384}$ denote the embedding of the LEED requirement for credit $c$. Semantic similarity is computed using cosine similarity:

\begin{equation}
S_{\mathrm{sem}}(c,q_k)
=
\frac{
\mathbf{e}(r_c)^{\top}\mathbf{e}(x_k)
}{
\left\|\mathbf{e}(r_c)\right\|_2
\left\|\mathbf{e}(x_k)\right\|_2
}.
\label{eq:semantic_score}
\end{equation}

The initial retrieval ranking is lexical-first. Semantic search is used when keyword retrieval produces too few candidate chunks, when the maximum lexical score is below a credit-specific sufficiency threshold, or when the model reports missing evidence with low confidence. This ordering is intentional. In this corpus, exact technical terms are usually stronger signals than generalized semantic similarity.

The final evidence set contains the top $K$ ranked chunks, where $K=8$ for the main experiments, subject to an 8,000-character evidence budget. Let $\mathrm{TopK}(\cdot)$ return the highest scoring chunks after de-duplication. The evidence set is

\begin{equation}
E_{b,c}^{(0)}
=
\mathrm{TopK}
\left(
\mathcal{Q}_{g(c)},
S_{\mathrm{lex}},
S_{\mathrm{sem}},
K
\right).
\label{eq:evidence_set}
\end{equation}

The superscript $(0)$ denotes the first retrieval attempt. The system supports one additional retrieval iteration. If the language model returns \textsc{insufficient\_data} with confidence below 0.9, it generates or selects additional search terms using the model's reported missing elements. The system may then search neighboring LEED categories or the full corpus. The revised evidence set is

\begin{equation}
E_{b,c}^{(1)}
=
E_{b,c}^{(0)}
\cup
\mathrm{Retrieve}
\left(
\mathcal{D}^{(b)},
W_c \cup W_c^{\mathrm{refined}}
\right).
\label{eq:iterative_retrieval}
\end{equation}

The second retrieval pass is not used to manufacture evidence. It is used to distinguish between two different failure modes: evidence that is absent from the corpus and evidence that was present but missed by the initial retrieval query.

\begin{algorithm}[htbp]
\caption{Three-Tier Agentic Evidence Retrieval}
\label{alg:retrieval}
\begin{algorithmic}[1]
\Require Credit $c$, indexed corpus $\mathcal{D}^{(b)}$, chunk set $\mathcal{Q}$
\Ensure Evidence set $E_{b,c}$

\State Restrict candidate set to category $\mathcal{Q}_{g(c)}$
\State Retrieve lexical candidates using $W_c$ \Comment{via Equation~\ref{eq:lexical_score}}
\State Rank credit-aware chunks and retain top candidates

\If{fewer than $K$ useful chunks are available}
    \State Compute semantic similarity \Comment{via Equation~\ref{eq:semantic_score}}
    \State Add highest-ranked semantic candidates
\EndIf

\State Construct initial evidence set $E_{b,c}^{(0)}$
\State Obtain preliminary LLM result $z_{b,c}^{(0)}$

\If{$z_{b,c}^{(0)}.\mathrm{verdict} = \textsc{insufficient\_data}$ \textbf{and} $z_{b,c}^{(0)}.\mathrm{confidence} < 0.9$}
    \State Derive refined terms from $z_{b,c}^{(0)}.\mathrm{missing\_elements}$
    \State Expand search to neighboring categories or the full corpus
    \State Construct $E_{b,c}^{(1)}$ \Comment{via Equation~\ref{eq:iterative_retrieval}}
    \State \Return $E_{b,c}^{(1)}$
\EndIf

\State \Return $E_{b,c}^{(0)}$
\end{algorithmic}
\end{algorithm}

\subsection{Language-Model Evidence Verification}
\label{sec:verification}

The language-model component receives three inputs: the canonical LEED requirement, the retrieved evidence set, and a structured output schema. The requirement is retrieved from a LEED knowledge base containing the evaluated credit requirements, documentation expectations, and quantitative criteria. The evidence set includes chunk text together with source provenance. The prompt asks the model to assess whether the retrieved evidence supports the requirement, not whether the building is broadly sustainable or whether a scorecard claims that the credit was pursued.

The primary inference model is \texttt{gemma3:4b}, executed locally through Ollama. The model is run in JSON mode with a maximum generation length of 1,024 tokens. The primary neuro-symbolic configuration uses temperature 0.85. This temperature was selected to reduce excessive abstention while retaining structured output constraints. Because higher temperature also introduces stochastic variation, the final decision uses self-consistency voting across three independent runs.

For each inference run $r \in \{1,2,3\}$, the model returns a structured response

\begin{equation}
z_{b,c}^{(r)}
=
\left[
v_{b,c}^{(r)},
\gamma_{b,c}^{(r)},
j_{b,c}^{(r)},
Q_{b,c}^{(r)},
M_{b,c}^{(r)},
N_{b,c}^{(r)}
\right],
\label{eq:llm_output}
\end{equation}

where $v_{b,c}^{(r)}$ is the predicted verdict, $\gamma_{b,c}^{(r)} \in [0,1]$ is model confidence, $j_{b,c}^{(r)}$ is a justification, $Q_{b,c}^{(r)}$ is a set of evidence quotations, $M_{b,c}^{(r)}$ is a set of missing elements, and $N_{b,c}^{(r)}$ is a dictionary of extracted numerical values.

The parser normalizes common output variation before aggregation. It maps case variants and semantically equivalent labels, such as \texttt{true}, \texttt{satisfied}, or \texttt{compliant}, to \textsc{pass}. It maps \texttt{not met} and \texttt{noncompliant} to \textsc{fail}. It maps uncertainty labels to \textsc{insufficient\_data}. Confidence values are clamped to the interval $[0,1]$. Invalid JSON, incomplete fields, and malformed numeric structures are handled through defensive defaults rather than silently discarded.

The aggregated language-model verdict is the modal class across three runs:

\begin{equation}
v_{b,c}^{\mathrm{LLM}}
=
\operatorname*{mode}_{y \in \mathcal{Y}}
\sum_{r=1}^{3}
\mathbb{1}
\left[
v_{b,c}^{(r)} = y
\right],
\qquad
\mathcal{Y} =
\{
\textsc{pass},
\textsc{fail},
\textsc{insufficient\_data}
\}.
\label{eq:self_consistency}
\end{equation}

The confidence assigned to the aggregated verdict is the mean confidence among runs that produced the modal verdict:

\begin{equation}
\gamma_{b,c}^{\mathrm{LLM}}
=
\frac{
\sum_{r=1}^{3}
\gamma_{b,c}^{(r)}
\mathbb{1}
\left[
v_{b,c}^{(r)} =
v_{b,c}^{\mathrm{LLM}}
\right]
}{
\sum_{r=1}^{3}
\mathbb{1}
\left[
v_{b,c}^{(r)} =
v_{b,c}^{\mathrm{LLM}}
\right]
}.
\label{eq:aggregated_confidence}
\end{equation}

Evidence quotations and numerical values are merged across the runs supporting the modal verdict. This approach avoids treating a single stochastic response as authoritative while retaining traceability to the passages that motivated the decision.

The model prompt includes four safeguards. First, it distinguishes scorecard mentions from substantive evidence. A scorecard entry alone is not sufficient to establish credit compliance. Second, it requires direct quotations from the retrieved documentation. Third, it asks the model to state what evidence is absent when it returns \textsc{insufficient\_data}. Fourth, it requires the extraction of any potentially relevant numerical values, even if the model believes the evidence is incomplete. The last safeguard is necessary because the deterministic numeric checker may be able to evaluate a threshold even when the model's narrative confidence is low.

\subsection{Deterministic Numeric Checking}
\label{sec:numeric_checker}

The numeric checker is the symbolic component of the pipeline. It applies deterministic LEED v4.1 threshold rules after the language model has extracted candidate values from the documentation. The checker does not attempt to interpret arbitrary project narratives. It performs a narrower task: given a credit identifier, a set of extracted values, and, when needed, supporting text, it evaluates the relevant formula or point table exactly.

This design avoids assigning arithmetic responsibility to the language model. A language model may correctly identify a proposed energy-use intensity and a baseline energy-use intensity, then still apply the wrong threshold or infer the wrong point value. The symbolic checker cannot resolve missing evidence, but it can guarantee consistent treatment once the necessary values are available.

For a quantitative credit $c$, let $V_{b,c}$ be the set of normalized values extracted from the model response and regex fallback. The deterministic checker is

\begin{equation}
R_c(V_{b,c}, E_{b,c}) =
\begin{cases}
(\textsc{pass}, \gamma_R, \delta_R), & \text{if LEED rule for } c \text{ is satisfied}, \\
(\textsc{fail}, \gamma_R, \delta_R), & \text{if LEED rule for } c \text{ is violated}, \\
(\textsc{insufficient\_data}, 0, \delta_R), & \text{if required values cannot be established},
\end{cases}
\label{eq:rule_engine}
\end{equation}

where $\gamma_R$ is deterministic checker confidence and $\delta_R$ is a human-readable explanation of the threshold comparison. In the present implementation, determinate numeric checks receive $\gamma_R=1.0$ because the uncertainty lies in value extraction, not in rule execution. If extraction is incomplete, the checker returns \textsc{insufficient\_data} rather than inferring a value.

The checker first normalizes nested value objects. For example,

\begin{verbatim}
{"percent_improvement": {"value": 18.0, "unit": "%"}}
\end{verbatim}

is flattened to a scalar representation:

\begin{verbatim}
{"percent_improvement": 18.0}
\end{verbatim}

If the language model does not provide the required values, the checker applies credit-specific regular-expression fallback patterns to the retrieved evidence. The fallback is limited to explicit numeric statements and does not estimate values from vague language. For example, EA-p2 and EA-c1 patterns search for percentage improvement relative to a baseline or ASHRAE reference. WE-p2 and WE-c2 patterns search for stated water-use reduction percentages. MR-c5 patterns search for diversion percentages. EQ-c7 patterns search for spatial daylight autonomy values, and SS-c5 patterns search for roof or paving SRI values.

For energy-improvement credits, the system can calculate the percentage improvement from EUI values when a directly reported improvement percentage is unavailable:

\begin{equation}
I_{\mathrm{energy}}
=
\frac{
\mathrm{EUI}_{\mathrm{baseline}}
-
\mathrm{EUI}_{\mathrm{proposed}}
}{
\mathrm{EUI}_{\mathrm{baseline}}
}
\times 100.
\label{eq:energy_improvement}
\end{equation}

For example, if the baseline EUI is 51 and the proposed EUI is 42, the computed improvement is 17.65\%. This value is then compared with the applicable prerequisite threshold or the EA-c1 point table. The implementation does not round upward before comparison. It preserves the extracted or calculated precision and applies the threshold directly.

For point-based credits, the checker implements a step function. Let $\mathcal{T}_c = \{(a_j,p_j)\}_{j=1}^{J}$ be the ordered set of threshold-point pairs for credit $c$, where $a_j$ is the minimum achievement value and $p_j$ is the awarded point value. The deterministic point assignment is

\begin{equation}
\mathrm{Points}_c(v)
=
\max
\left\{
p_j :
v \geq a_j,\,
(a_j,p_j) \in \mathcal{T}_c
\right\},
\label{eq:point_assignment}
\end{equation}

with $\mathrm{Points}_c(v)=0$ when no threshold is met. This formulation is used for credits such as EA-c1, EA-c5, WE-c2, MR-c5, and EQ-c7.

Table~\ref{tab:numeric_checkers} lists the implemented credit-specific checkers. The table identifies the decision rule and required field set. A credit is routed to the checker only when the credit is in the implemented numeric set or when the language model flags it as requiring numerical verification.

\begin{table*}[htbp]
\centering
\small
\caption{Deterministic numeric checkers implemented in the neuro-symbolic pipeline.}
\label{tab:numeric_checkers}
\begin{tabular}{lll}
\toprule
\textbf{Credit} & \textbf{Deterministic Decision Rule} & \textbf{Required Value(s)} \\
\midrule
EA-p2 & PASS if energy improvement is at least 5\% & \texttt{percent\_improvement} \\
EA-c1 & Assign points through LEED energy-improvement threshold table & \texttt{percent\_improvement} \\
EA-c5 & 1\%=1 point, 5\%=2 points, 10\%=3 points & \texttt{renewable\_offset\_pct} \\
WE-p2 & PASS if indoor water reduction is at least 20\% & \texttt{water\_reduction\_pct} \\
WE-c2 & Assign points from 25\% through 50\% reduction thresholds & \texttt{water\_reduction\_pct} \\
MR-c5 & 50\%=1 point; 75\%=2 points & \texttt{waste\_diversion\_pct} \\
EQ-p1 & PASS if compliance with ASHRAE 62.1 is documented & \texttt{ventilation\_reference} \\
EQ-c7 & Evaluate sDA threshold and associated point level & \texttt{sda\_pct} \\
EQ-c8 & PASS if qualifying views cover at least 75\% of occupied area & \texttt{views\_pct} \\
SS-c3 & PASS if open space covers at least 30\% of site area & \texttt{open\_space\_pct} \\
SS-c5 & Evaluate roof SRI, paving SRI, and qualifying coverage & \texttt{roof\_sri}, \texttt{paving\_sri}, \texttt{coverage\_pct} \\
LT-c5 & Check short-term and long-term bicycle storage thresholds & \texttt{short\_term\_bike\_pct}, \texttt{long\_term\_bike\_pct} \\
\bottomrule
\end{tabular}
\end{table*}

Checker integration follows a non-destructive override rule. The checker may replace the prior result only when it returns \textsc{pass} or \textsc{fail}. When it returns \textsc{insufficient\_data}, the system preserves the language-model result. This policy is intentional. The checker should correct arithmetic and threshold errors, not introduce a new negative decision merely because its narrow extractor failed to recover a value that the model interpreted from context.

\begin{algorithm}[htbp]
\caption{Deterministic Numeric Check and Controlled Override}
\label{alg:numeric_checker}
\begin{algorithmic}[1]
\Require Credit $c$, LLM result $(v_L,\gamma_L)$, extracted values $N$, evidence $E$
\Ensure Updated result $(v_F,\gamma_F,\delta_F)$

\If{no deterministic checker exists for $c$}
    \State \Return $(v_L,\gamma_L,\varnothing)$
\EndIf

\State $V \gets \mathrm{FlattenNestedValues}(N)$
\If{required fields for $c$ are absent from $V$}
    \State $V \gets \mathrm{RegexExtract}(E,c)$
\EndIf

\State $(v_R,\gamma_R,\delta_R) \gets R_c(V,E)$ \Comment{via Equation~\ref{eq:rule_engine}}

\If{$v_R \in \{\textsc{pass},\textsc{fail}\}$}
    \State $v_F \gets v_R$
    \State $\gamma_F \gets \max(\gamma_L,\gamma_R)$
    \State $\delta_F \gets \delta_R$
\Else
    \State $v_F \gets v_L$
    \State $\gamma_F \gets \gamma_L$
    \State $\delta_F \gets$ ``No deterministic override: insufficient values''
\EndIf

\State \Return $(v_F,\gamma_F,\delta_F)$
\end{algorithmic}
\end{algorithm}

\subsection{Promotion of Conservative Uncertainty Verdicts}
\label{sec:promotion}

Language models frequently return \textsc{insufficient\_data} when documentation is present but distributed across several partial pieces of evidence. This behavior can be useful when the corpus genuinely lacks the required materials. It becomes harmful when the project corpus is documentation-rich and most attempted credits are supported. The problem is a calibration problem. A high-confidence uncertainty verdict should not be treated identically across a project where 80\% of credits are well documented and a project where only 40\% are documented.

The system therefore includes a conservative-to-aggressive promotion step. The step is applied after language-model aggregation and before deterministic numeric checking. If the model returns \textsc{insufficient\_data} with confidence above a calibrated threshold $\tau_b$, the provisional result is promoted to \textsc{pass}. Formally, the promotion function is

\begin{equation}
P(v,\gamma;\tau_b) =
\begin{cases}
\textsc{pass}, &
\text{if } v=\textsc{insufficient\_data} \land \gamma > \tau_b, \\
v, & \text{otherwise}.
\end{cases}
\label{eq:promotion_rule}
\end{equation}

Promotion is not intended to imply that uncertainty is evidence of compliance. It is a calibration intervention for a known directional bias in the model. The rule is appropriate only if $\tau_b$ is selected without examining the labels of the final evaluation set.

To avoid test-set leakage, the threshold must be selected using development data only. In a leave-one-building-out protocol, for each test building $b$, the threshold is selected on the remaining buildings $\mathcal{B}\setminus\{b\}$:

\begin{equation}
\tau_b^{*}
=
\operatorname*{arg\,max}_{\tau \in \mathcal{T}}
\frac{
1
}{
\left|
\mathcal{Y}_{\mathrm{dev}}
\right|
}
\sum_{(i,c)\in\mathcal{Y}_{\mathrm{dev}}}
\mathbb{1}
\left[
P\left(
v_{i,c}^{\mathrm{LLM}},
\gamma_{i,c}^{\mathrm{LLM}};
\tau
\right)
=
y_{i,c}
\right],
\label{eq:threshold_calibration}
\end{equation}

where $\mathcal{T}$ is a pre-specified threshold grid, for example $\{0.50,0.55,\dots,0.90\}$, $\mathcal{Y}_{\mathrm{dev}}$ is the development set, and $y_{i,c}$ is the expert ground-truth label. The reported value of $\tau=0.6$ should therefore be described as either a development-set optimum or a global fixed threshold, depending on the actual experimental protocol. If it was selected after observing performance on all four buildings, it should be treated as exploratory rather than confirmatory.

Promotion occurs before numeric checking. This ordering gives the symbolic module priority whenever it has sufficient numerical information. If the checker establishes PASS or FAIL, it overrides the promoted provisional verdict. If the checker cannot establish a determination, the promoted result remains. The final decision is therefore controlled by a hierarchy: deterministic evidence dominates when available, while calibrated model judgment governs only where deterministic evaluation cannot be completed.

\subsection{Final Output and Audit Trail}
\label{sec:output}

The final output is a structured \texttt{VerificationResult} object designed for both automated evaluation and human review. Each result includes the canonical credit identifier, final verdict, final confidence, language-model verdict before promotion, post-promotion verdict, numeric-checker result if invoked, retrieved evidence quotations, source files, page references, numerical values, missing evidence elements, and a concise rationale.

The system records the full decision path rather than only the final class label. This is important because certification review requires accountability. A reviewer must be able to distinguish three situations: a credit that passed because the language model found clear narrative evidence, a credit that passed because a deterministic threshold was satisfied, and a credit that remained uncertain because the required evidence could not be located. The resulting audit trail also supports error analysis. It allows investigators to determine whether an incorrect verdict arose from retrieval failure, extraction failure, language-model reasoning, numeric parsing, threshold logic, or calibration.

\section{Experimental Setup}
\label{sec:experiments}

\subsection{Study Corpus and Ground Truth}
\label{sec:dataset}

The evaluation corpus for this study comprises documentation from four University of Texas at Austin projects: Energy Engineering Building (Building 1), Auditorium (Building 2), Administrative Building 1 (Building 3), and Administrative Building 2 (Building 4). The corpus contains 484 PDFs and 153 credit-level ground-truth decisions. Each ground-truth decision was verified by a LEED Accredited Professional using project documentation and official GBCI review information.

The corpus is deliberately challenging because buildings differ in documentation density, folder organization, file naming, certification history, and credit distribution. Energy Engineering Building and Building 2 are PASS-heavy projects. Building 3 has a substantially larger proportion of \textsc{insufficient\_data} decisions. Building 4 is more balanced. These differences enable evaluation not only of overall accuracy but also of calibration under changing class distributions.

\begin{table}[H]
\centering
\small
\caption{Building corpus and credit-level ground-truth distribution.}
\label{tab:building_stats}
\begin{tabular}{lccccc}
\toprule
\textbf{Building} & \textbf{Credits} & \textbf{PDFs} & \textbf{PASS} & \textbf{INSUFF.} & \textbf{PASS Rate} \\
\midrule
Building 1 & 49 & 126 & 40 & 9 & 81.6\% \\
Building 2 & 34 & 168 & 28 & 6 & 82.4\% \\
Building 3 & 35 & 111 & 15 & 20 & 42.9\% \\
Building 4 & 35 & 79 & 19 & 16 & 54.3\% \\
\midrule
Total & 153 & 484 & 102 & 51 & 66.7\% \\
\bottomrule
\end{tabular}
\end{table}

\subsection{Models and Experimental Conditions}
\label{sec:models}

Five locally deployed models are evaluated: \texttt{gemma3:4b}, \texttt{llama3.1:8b}, \texttt{llava:7b}, \texttt{moondream}, and \texttt{smollm3:3b}. All inference is performed locally through Ollama. No project documents are transmitted to an external model provider. This design supports privacy-sensitive deployment in institutional and commercial certification contexts.

The main baseline is text-only \texttt{llama3.1:8b} with three-run self-consistency. The primary text-only model is \texttt{gemma3:4b}. The neuro-symbolic condition uses \texttt{gemma3:4b}, credit-aware retrieval, three-run self-consistency, promotion logic, and the deterministic numeric checker. Vision-language conditions use \texttt{gemma3:4b} with up to four document images rendered at 150 DPI. Additional vision experiments use \texttt{llava:7b} and \texttt{moondream} at 150 and 300 DPI. Prompt experiments evaluate Baseline, Chain-of-Thought, Few-shot, and Rubric prompt forms.

\subsection{Evaluation Metrics}
\label{sec:metrics}

The primary metric is credit-level accuracy:

\begin{equation}
\mathrm{Accuracy}
=
\frac{
1
}{
|\mathcal{Y}|
}
\sum_{(b,c)\in\mathcal{Y}}
\mathbb{1}
\left[
\hat{y}_{b,c} = y_{b,c}
\right],
\label{eq:accuracy}
\end{equation}

where $\mathcal{Y}$ is the evaluated set of building-credit pairs, $\hat{y}_{b,c}$ is the system verdict, and $y_{b,c}$ is expert ground truth.

Accuracy is reported overall, by building, by LEED category, by credit, and by experimental condition. Because the corpus contains PASS and INSUFFICIENT\_DATA ground truth but no verified FAIL cases, class-specific analysis focuses on the distinction between supported compliance and insufficient evidence. The study also reports confusion matrices, numeric-checker invocation rate, numeric-checker override rate, promotion recovery count, and prompt-specific accuracy deltas.

For prompt comparison, the change relative to the baseline prompt is calculated as

\begin{equation}
\Delta A_m^{(b)}
=
A_m^{(b)}
-
A_{\mathrm{baseline}}^{(b)},
\label{eq:prompt_delta}
\end{equation}

where $A_m^{(b)}$ is accuracy for prompt mode $m$ on building $b$. This metric reveals whether a prompt change improves performance consistently or merely shifts the decision bias in a way that happens to fit one building's class distribution.

The primary statistical interpretation is descriptive because the corpus is limited to four buildings and 153 credit decisions. The paper therefore reports exact counts alongside percentages, avoids claims of population-level generalization, and treats building-level variation as a central result rather than noise to be averaged away.

\section{Results}
\label{sec:results}

This section reports the empirical performance of the evaluated language models and the proposed neuro-symbolic pipeline across four LEED-certified buildings. Results are organized from aggregate model behavior to category-level performance, deterministic numeric verification, promotion logic, multimodal ablations, prompt sensitivity, and computational cost. Accuracy is reported as the proportion of credit-level verdicts that matched the validated ground truth for each building. Unless otherwise stated, results are descriptive because the limited number of buildings and credit-level observations does not support strong claims of statistical generalizability beyond the present corpus.

\subsection{Overall Model Comparison}

\begin{figure}[htbp]
\centering
\includegraphics[width=\columnwidth]{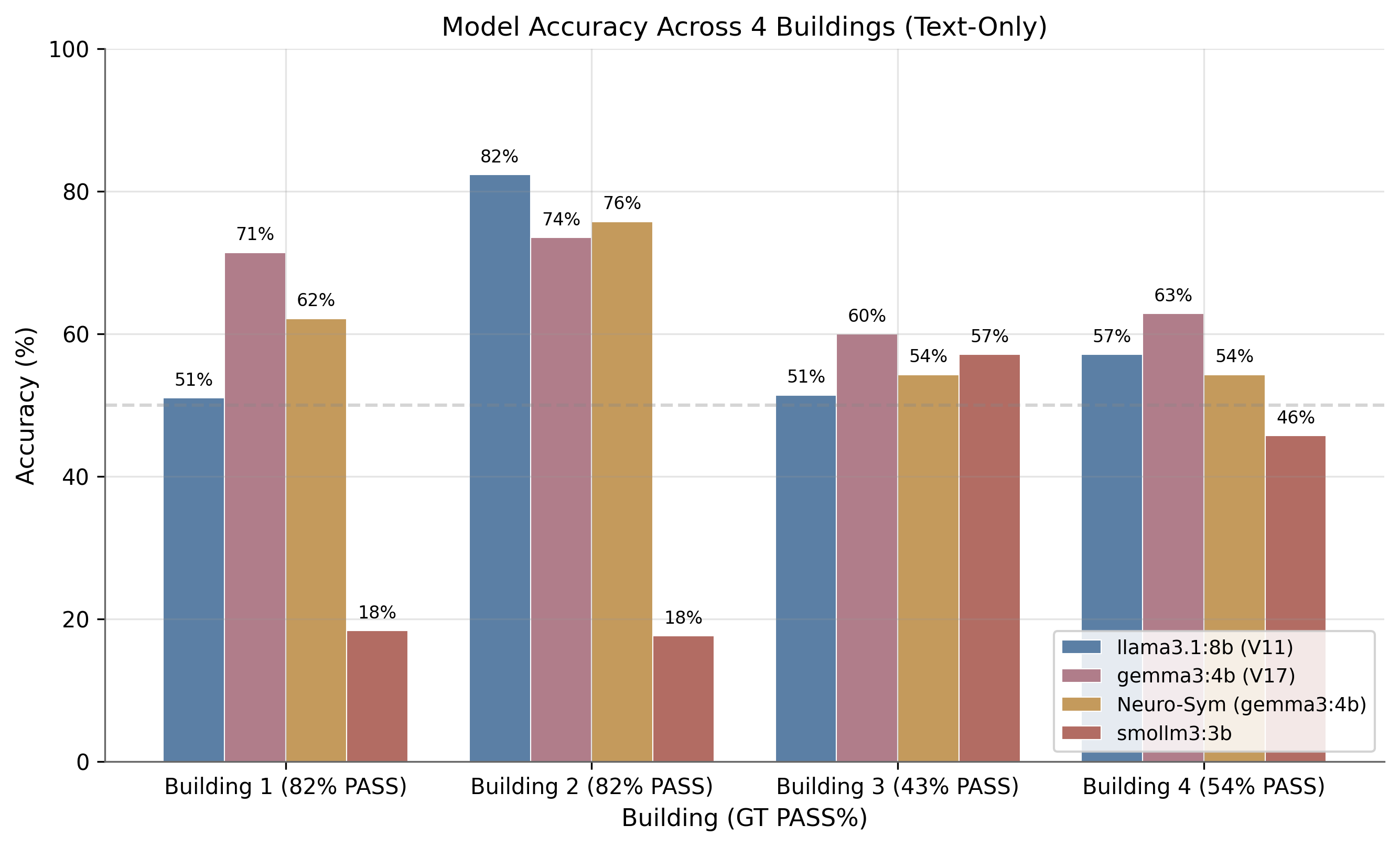}
\caption{Text-only model accuracy across four buildings. gemma3:4b achieves the highest overall accuracy (67.3\%), outperforming llama3.1:8b by 6.8~pp despite having half the parameters. smollm3:3b performs poorly on PASS-heavy buildings due to conservative bias.}
\label{fig:model_accuracy}
\end{figure}

Table~\ref{tab:overall_accuracy} presents text-only credit-verification accuracy across all evaluated models and buildings, with corresponding group comparisons shown in Figure~\ref{fig:model_accuracy}. The results show that model size alone did not determine performance. The 4B-parameter gemma3:4b model achieved the strongest overall text-only performance, reaching 67.3\% accuracy across the full evaluation set. This result exceeded the 60.5\% accuracy of llama3.1:8b by 6.8 percentage points, despite gemma3:4b having approximately half as many parameters. The finding is important because it indicates that a smaller, locally deployable model can be better calibrated for evidence-grounded compliance judgments than a larger general-purpose language model.

\begin{table}[H]
\centering
\small
\caption{Text-only accuracy across four buildings. Accuracy denotes the percentage of credit-level verdicts matching the validated ground truth.}
\label{tab:overall_accuracy}
\begin{tabular}{lccccc}
\toprule
\textbf{Model} & \textbf{Building 1} & \textbf{Building 2} & \textbf{Building 3} & \textbf{Building 4} & \textbf{Overall} \\
\midrule
llama3.1:8b (V11)   & 51.0\% & 82.4\% & 51.4\% & 57.1\% & 60.5\% \\
gemma3:4b (V17)     & 71.4\% & 73.5\% & 60.0\% & 62.9\% & \textbf{67.3\%} \\
Neuro-Sym (gemma3)  & 62.2\% & 73.5\% & 54.3\% & 54.3\% & 61.6\% \\
smollm3:3b          & 18.4\% & 17.6\% & 57.1\% & 45.7\% & 34.7\% \\
\bottomrule
\end{tabular}
\end{table}

gemma3:4b was the best-performing text-only model on three of the four buildings: Building 1, Building 3, and Building 4. Its advantage was particularly clear on Building 1, where it achieved 71.4\% accuracy compared with 51.0\% for llama3.1:8b. Building 1 contains a high proportion of PASS credits and dense evidence distributed across documentation sets. The stronger result for gemma3:4b suggests that the model more consistently recognized sufficient evidence in these documents rather than defaulting to \texttt{INSUFFICIENT\_DATA}. On Building 3 and Building 4, gemma3:4b also improved on llama3.1:8b by 8.6 and 5.8 percentage points, respectively. These gains are modest in absolute terms but meaningful in a credit-level verification task, where each misclassification may affect project teams' review priorities.

Building 2 produced a different pattern. llama3.1:8b achieved its highest building-specific accuracy on Building 2 at 82.4\%, exceeding gemma3:4b by 8.9 percentage points. This result suggests that Building 2's evidence corpus may have been comparatively easier for the larger model to interpret. Building 2 had a high proportion of PASS credits and relatively well-structured documentation, which may have reduced the ambiguity that normally causes larger general-purpose models to hedge or overinterpret evidence. The result also shows that gemma3:4b was not uniformly superior. Model choice remained sensitive to document organization, evidence density, and the distribution of ground-truth verdicts.

The neuro-symbolic configuration achieved 61.6\% overall accuracy. It therefore trailed gemma3:4b text-only by 5.7 percentage points, even though it introduced a deterministic numeric checker and promotion logic intended to improve reliability. This aggregate result should not be interpreted as a failure of the neuro-symbolic approach. Instead, it reveals an important asymmetry. The symbolic layer substantially improved performance on credits governed by explicit quantitative thresholds, especially in Energy and Atmosphere. However, it did not improve, and sometimes reduced performance, for qualitative credits where compliance depends on document interpretation rather than numerical comparison. The aggregate score therefore conceals strong gains on a small set of high-value quantitative credits and losses across several qualitative categories.

The smallest evaluated model, smollm3:3b, performed poorly overall at 34.7\%. Its behavior was strongly conservative. In many cases, it returned \texttt{INSUFFICIENT\_DATA} despite the presence of evidence that supported a PASS decision. This pattern was especially pronounced on Building 1 and Building 2, where accuracy fell below 20\%. The result indicates that model compactness alone is not sufficient for local deployment in this task. A small model can be computationally efficient, but it must also reliably distinguish absent evidence from evidence that is present but distributed across several documents.

The building-level pattern also reveals that task difficulty was not uniform. Building 2 was the easiest building for the larger models, with all non-small-model configurations reaching at least 73.5\% accuracy. Building 1 and Building 3 were more difficult. Building 1 challenged models because it contained a large, PASS-heavy corpus in which the main risk was false-negative \texttt{INSUFFICIENT\_DATA} classification. Building 3 posed a different challenge because its lower PASS rate required the models to discriminate more carefully between complete evidence and incomplete documentation. These differences motivate the subsequent analyses of category-specific performance, promotion behavior, and prompt sensitivity.

\subsection{Neuro-Symbolic Performance by LEED Category}

\begin{figure}[htbp]
\centering
\includegraphics[width=\columnwidth]{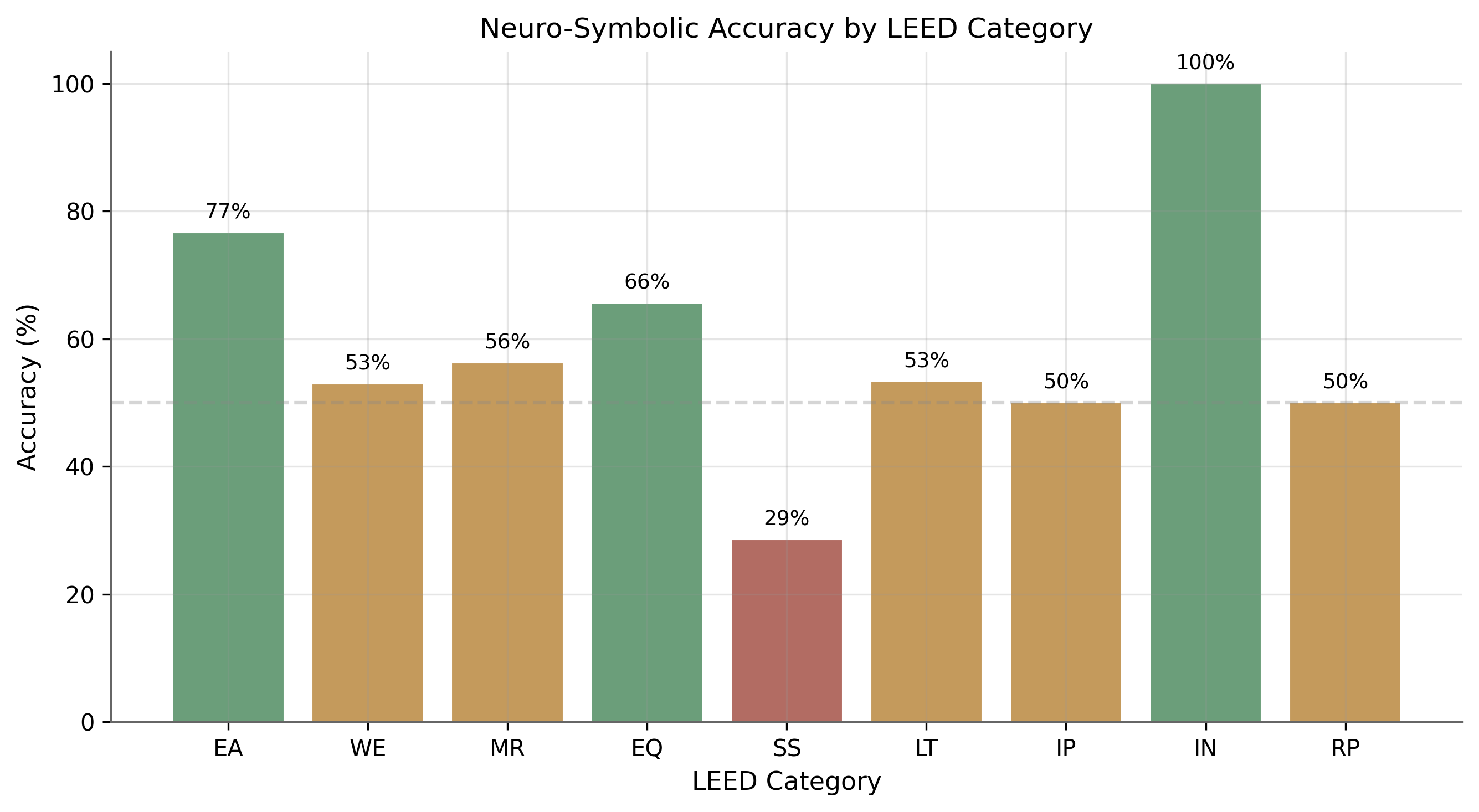}
\caption{Neuro-symbolic accuracy by LEED category across all four buildings. EA achieves the highest accuracy (76.7\%) due to numeric checker intervention. SS shows the largest degradation ($-$45.1~pp vs.\ text-only), revealing that Sustainable Sites evidence is poorly captured by text-based retrieval.}
\label{fig:ns_by_category}
\end{figure}

Table~\ref{tab:ns_category} compares neuro-symbolic and text-only gemma3:4b performance by LEED category, with per-category accuracy visualized in Figure~\ref{fig:ns_by_category}. The results demonstrate that the neuro-symbolic architecture was not uniformly beneficial. Its effect depended on whether the credit included a deterministic quantitative rule that could be delegated to the numeric checker.

\begin{table}[htbp]
\centering
\small
\caption{Neuro-symbolic accuracy by LEED category across all buildings. Delta represents the neuro-symbolic accuracy minus the V17 text-only accuracy.}
\label{tab:ns_category}
\begin{tabular}{lcccc}
\toprule
\textbf{Category} & \textbf{Neuro-Sym} & \textbf{V17 Text} & \textbf{Delta} & \textbf{Quant. Credits} \\
\midrule
EA  & 76.7\% & 57.6\% & +19.1 pp & 3 \\
WE  & 52.9\% & 52.9\% & +0.0 pp  & 2 \\
MR  & 56.2\% & 77.8\% & -21.5 pp & 1 \\
EQ  & 65.6\% & 74.3\% & -8.7 pp & 3 \\
SS  & 28.6\% & 73.7\% & -45.1 pp & 2 \\
LT  & 53.3\% & 60.0\% & -6.7 pp  & 1 \\
IP  & 50.0\% & 75.0\% & -25.0 pp & 0 \\
IN  & 100.0\% & 75.0\% & +25.0 pp & 0 \\
RP  & 50.0\% & 75.0\% & -25.0 pp & 0 \\
\bottomrule
\end{tabular}
\end{table}

Energy and Atmosphere produced the clearest neuro-symbolic gain. Accuracy increased from 57.6\% under text-only gemma3:4b to 76.7\% under the neuro-symbolic configuration, a gain of 19.1 percentage points. This result is consistent with the structure of EA credits. Several EA requirements depend on explicit numerical thresholds, including energy improvement over a baseline, renewable-energy offset, and energy-performance point tables. These requirements are well suited to a division of labor in which the LLM identifies candidate values and the deterministic checker applies the relevant rule. The checker removes arithmetic evaluation from the language model, reducing the risk that a plausible but incorrect generated statement determines the final verdict.

The Water Efficiency category showed no aggregate change. Both systems achieved 52.9\% accuracy. This neutral result is informative because it shows that the presence of quantitative requirements is not, by itself, sufficient to ensure improvement. Water credits require reliable identification of fixture-flow values, baseline assumptions, and aggregate water reduction calculations. When those values were not extracted accurately, the deterministic component could not provide a valid correction. The absence of improvement in WE therefore reflects an upstream extraction limitation rather than a weakness in threshold comparison itself.

The largest decline occurred in Sustainable Sites. Neuro-symbolic accuracy was 28.6\%, compared with 73.7\% for text-only gemma3:4b, a decrease of 45.1 percentage points. Sustainable Sites evidence often appears in civil plans, site drawings, heat-island calculations, and land-use narratives. Such evidence is less likely to be captured by text-only retrieval and less likely to be represented by a single, clearly labeled numerical field. For example, open-space verification may require a site-area calculation embedded in a plan set, while heat-island verification may require linking roof and paving material properties to their respective surface areas. The numeric checker can only evaluate values that the retrieval and extraction stages surface correctly. When that evidence was incomplete or poorly extracted, the symbolic layer could not compensate for the missing context.

Materials and Resources, Integrative Process, and Regional Priority also declined under the neuro-symbolic configuration. These categories depend largely on qualitative or documentary evidence such as product declarations, life-cycle considerations, integrative-process meeting records, and region-specific credit status. They do not reduce naturally to a small set of deterministic thresholds. For these categories, the additional pipeline stages introduced opportunities for retrieval mismatches and confidence-based promotion effects without providing a corresponding symbolic advantage. The result does not imply that neuro-symbolic methods are inappropriate for qualitative credits. It instead indicates that a numeric checker alone is an incomplete symbolic representation for a rating system that includes both numeric and narrative requirements.

Innovation was the only non-quantitative category that improved. Neuro-symbolic accuracy reached 100.0\%, compared with 75.0\% for text-only gemma3:4b. This result should be interpreted cautiously because the category contained few observations. Innovation credits frequently have distinctive documentation, such as an innovation narrative, pilot-credit template, or exemplary-performance statement. These recognizable patterns may have made the evidence easier to retrieve and classify, while the promotion mechanism may have corrected otherwise conservative \texttt{INSUFFICIENT\_DATA} predictions. The small sample size means that this apparent advantage should be treated as suggestive rather than conclusive.

Overall, the category analysis identifies the central boundary condition of the proposed architecture. The neuro-symbolic pipeline is most useful where compliance depends on a deterministic quantitative rule and the necessary values can be extracted from documentation. It is less effective where compliance requires spatial interpretation, cross-document synthesis, or qualitative assessment. This boundary condition guides the credit-level analysis that follows.

\subsection{Numeric Checker Effectiveness}

\begin{figure}[htbp]
\centering
\includegraphics[width=\columnwidth]{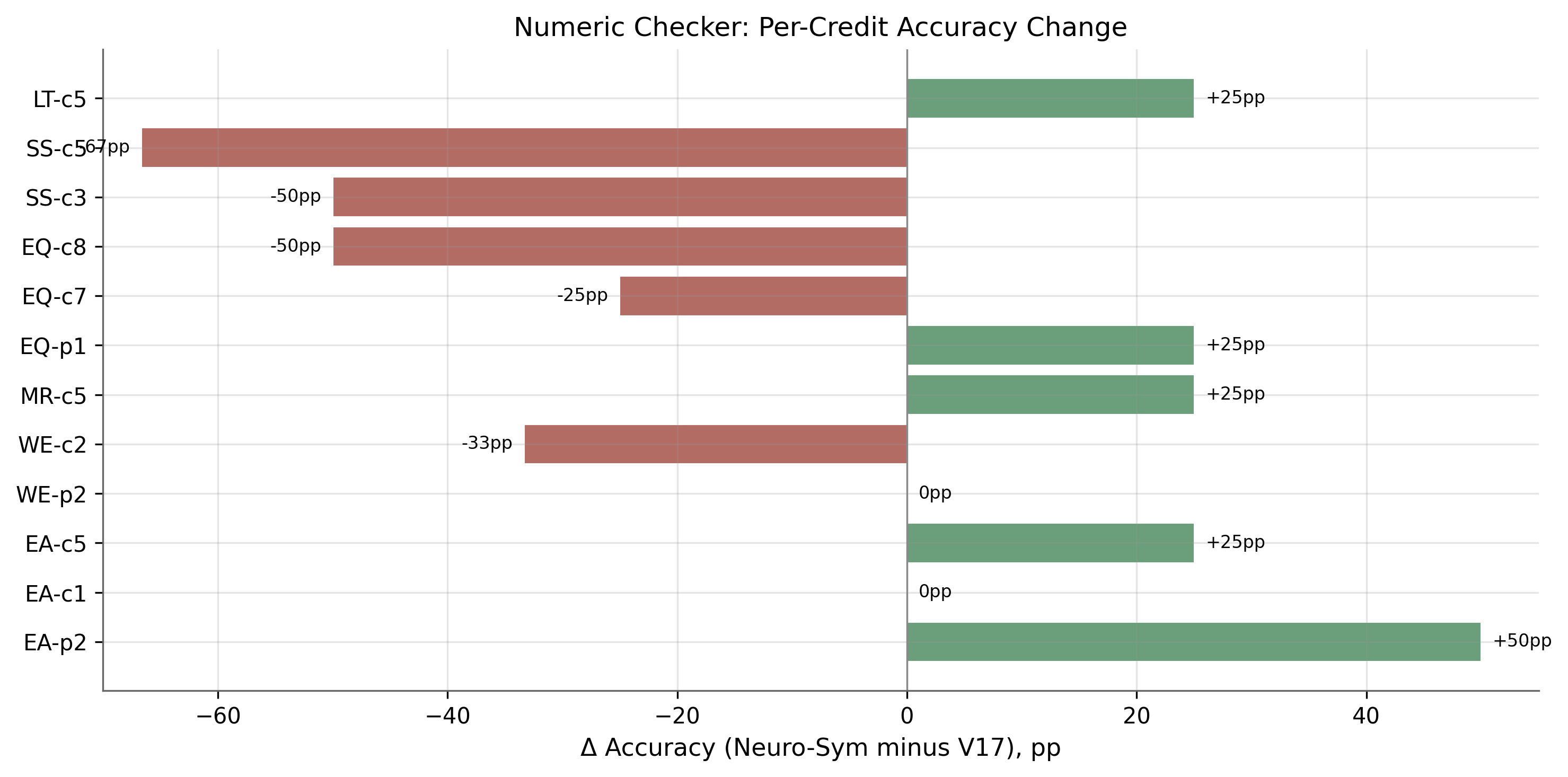}
\caption{Per-credit accuracy delta (neuro-symbolic minus V17 text-only) across 12 quantitative credits. EA-p2 shows the largest improvement (+50~pp, 50\%$\rightarrow$100\%). Regressions on SS-c5, SS-c3, EQ-c8, and WE-c2 are caused by extraction failures, not checker logic errors.}
\label{fig:numeric_checker_delta}
\end{figure}

Table~\ref{tab:numeric_results} isolates the performance of the deterministic numeric checker on the quantitative credits supported by the system, and Figure~\ref{fig:numeric_checker_delta} visualizes per-credit accuracy changes. The results show that the checker produced substantial improvements on several credits, but its benefit depended on successful extraction of the values required by each LEED rule.

\begin{table}[htbp]
\centering
\small
\caption{Numeric checker effectiveness on quantitative credits. ``Checker invoked'' denotes the number of buildings for which the checker received an applicable credit evaluation.}
\label{tab:numeric_results}
\begin{tabular}{lcccc}
\toprule
\textbf{Credit} & \textbf{V17 Text} & \textbf{Neuro-Sym} & \textbf{Delta} & \textbf{Checker Invoked} \\
\midrule
EA-p2 & 50.0\% & 100.0\% & +50.0 pp & 4/4 buildings \\
EA-c1 & 75.0\% & 75.0\% & +0.0 pp  & 4/4 \\
EA-c5 & 0.0\%  & 25.0\%  & +25.0 pp & 3/4 \\
WE-p2 & 50.0\% & 50.0\% & +0.0 pp  & 4/4 \\
WE-c2 & 33.3\% & 0.0\%   & -33.3 pp & 3/4 \\
MR-c5 & 50.0\% & 75.0\% & +25.0 pp & 3/4 \\
EQ-p1 & 75.0\% & 100.0\% & +25.0 pp & 2/4 \\
EQ-c7 & 75.0\% & 50.0\%  & -25.0 pp & 3/4 \\
EQ-c8 & 75.0\% & 25.0\%  & -50.0 pp & 3/4 \\
SS-c3 & 50.0\% & 0.0\%   & -50.0 pp & 2/4 \\
SS-c5 & 100.0\% & 33.3\% & -66.7 pp & 2/4 \\
LT-c5 & 50.0\% & 75.0\% & +25.0 pp & 2/4 \\
\bottomrule
\end{tabular}
\end{table}

EA-p2, Minimum Energy Performance, was the clearest success. Text-only gemma3:4b achieved 50.0\% accuracy, whereas the neuro-symbolic pipeline achieved 100.0\% accuracy across all four buildings. EA-p2 requires a project to demonstrate a minimum percentage improvement relative to an energy-code baseline. This structure is well matched to deterministic verification. Once proposed and baseline values, or a directly stated percentage improvement, were identified, the checker applied the LEED threshold of at least 5\% improvement without relying on the language model to perform arithmetic. The 50.0-percentage-point increase is especially consequential because EA-p2 is a prerequisite. An error in a prerequisite assessment has a different practical significance from an error in a discretionary point-bearing credit.

The checker also improved EA-c5, MR-c5, EQ-p1, and LT-c5 by 25.0 percentage points each. For EA-c5, the system applied the renewable-energy offset scale after identifying the relevant offset percentage. For MR-c5, it evaluated construction-waste diversion percentages against the applicable thresholds. For EQ-p1, it detected documentary reference to the required ventilation standard, ASHRAE 62.1. For LT-c5, it assessed bicycle parking capacity relative to building occupancy. These credits vary in their domain content, but they share a favorable computational structure: the compliance condition can be represented as a relatively compact rule, and the relevant evidence is often expressed as a number or a recognizable standard reference.

EA-c1 and WE-p2 did not improve. In EA-c1, both systems reached 75.0\% accuracy. The absence of a gain indicates that the text-only model was already correct in most cases or that the values required for point determination were not consistently extracted. In WE-p2, the checker did not change the aggregate outcome because the verification requires more than a single number. It depends on fixture-level flow values, baseline assumptions, and an aggregate reduction calculation. A deterministic checker can evaluate a reported reduction percentage, but it cannot independently reconstruct a complete water-use model unless the document parser reliably extracts the full fixture schedule and its associated assumptions.

Several credits declined under the neuro-symbolic configuration. The most substantial declines occurred in SS-c5, SS-c3, EQ-c8, and WE-c2. These declines should not be interpreted as evidence that deterministic checking is intrinsically harmful. The observed pattern is more specific. The checker requires correctly typed and correctly contextualized values. When the extraction stage returns no value, the checker preserves the underlying LLM result. When it returns an incorrect or ambiguously typed value, the checker may evaluate the wrong quantity. WE-c2 illustrates this failure mode. In some cases, values associated with airflow or other building-system parameters were incorrectly treated as water-reduction evidence. The resulting error was an extraction and schema-alignment problem, not a threshold-comparison error.

EQ-c7, EQ-c8, SS-c3, and SS-c5 exposed a second limitation. These credits require values that are often distributed across drawings, simulations, schedules, or multiple project narratives. Daylight autonomy, quality views, open-space area, roof reflectance, paving reflectance, and surface coverage may not appear as a single explicit value in a text-extracted document. When the pipeline could not retrieve or extract the necessary value, the checker returned \texttt{INSUFFICIENT\_DATA}. The system was designed so that an uncertain checker output would not automatically override a language-model verdict. Nevertheless, incomplete value extraction limited the checker's ability to correct the original decision.

Across the evaluation set, the numeric checker was invoked on 41 of 141 applicable credit evaluations, or 29\%. Of those invocations, 24 returned a determinate PASS or FAIL result and therefore supplied an explicit rule-based verdict. The remaining 17 returned \texttt{INSUFFICIENT\_DATA}, indicating that the checker did not receive sufficient structured input to evaluate the applicable rule. This outcome is informative. The numeric engine itself is not the primary bottleneck. The bottleneck is reliable extraction of semantically correct values from heterogeneous construction documents. In other words, deterministic verification can be exact once it receives valid inputs, but it cannot repair evidence that retrieval or extraction failed to surface.

\subsection{Promotion Logic Effectiveness}

The pipeline included a promotion rule intended to address conservative \texttt{INSUFFICIENT\_DATA} behavior. Under this rule, an \texttt{INSUFFICIENT\_DATA} verdict with confidence above $\tau = 0.6$ was eligible for promotion to PASS before deterministic checking. Table~\ref{tab:promotion} reports the change in building-level accuracy before and after this operation.

\begin{table}[htbp]
\centering
\small
\caption{Effect of confidence-based promotion using $\tau=0.6$. ``Promoted'' denotes the number of \texttt{INSUFFICIENT\_DATA} outputs changed to PASS; it does not imply that every promoted verdict was correct.}
\label{tab:promotion}
\begin{tabular}{lcccc}
\toprule
\textbf{Building} & \textbf{Before Promotion} & \textbf{After Promotion} & \textbf{Delta} & \textbf{Promoted} \\
\midrule
Building 1 & 51.4\% & 62.2\% & +10.8 pp & 11/38 \\
Building 2 & 67.6\% & 73.5\% & +5.9 pp  & 7/33 \\
Building 3 & 51.4\% & 54.3\% & +2.9 pp  & 4/35 \\
Building 4 & 51.4\% & 54.3\% & +2.9 pp  & 4/35 \\
\bottomrule
\end{tabular}
\end{table}

Promotion improved accuracy for all four buildings, but its magnitude varied substantially. Building 1 experienced the largest improvement, increasing from 51.4\% to 62.2\%, or 10.8 percentage points. Building 2 improved by 5.9 percentage points. Building 3 and Building 4 each improved by 2.9 percentage points. The pattern follows the class distribution of the building-level ground truth. Building 1 and Building 2 were PASS-heavy, meaning that a high-confidence \texttt{INSUFFICIENT\_DATA} verdict was more likely to be a false negative than a correct rejection. Building 3 and Building 4 had lower PASS rates, so the same verdict type was more often appropriate.

The promotion rule changed 26 verdicts across the four buildings: 11 in Building 1, seven in Building 2, four in Building 3, and four in Building 4. These 26 cases should be described as promoted predictions, not as 26 recovered false negatives. The net accuracy gains in Table~\ref{tab:promotion} show that promotion corrected more false negatives than it introduced false positives, especially in Building 1 and Building 2. However, the difference between the number of promotions and the net number of corrected predictions indicates that not all promoted outcomes were beneficial. This is expected. A fixed confidence threshold is a calibration mechanism, not a proof of correctness.

The results show that confidence-based promotion can be useful when the system is deployed on a project with a high expected PASS rate. It should not be treated as a universal default. On a building where many credits genuinely lack evidence, a low promotion threshold could convert valid uncertainty into false PASS decisions. The observed building-level variation therefore supports using promotion as a configurable mechanism that should be calibrated to project type, documentation maturity, and expected credit-attempt profile.

\subsection{Multimodal Ablation}

\begin{figure}[htbp]
\centering
\includegraphics[width=\columnwidth]{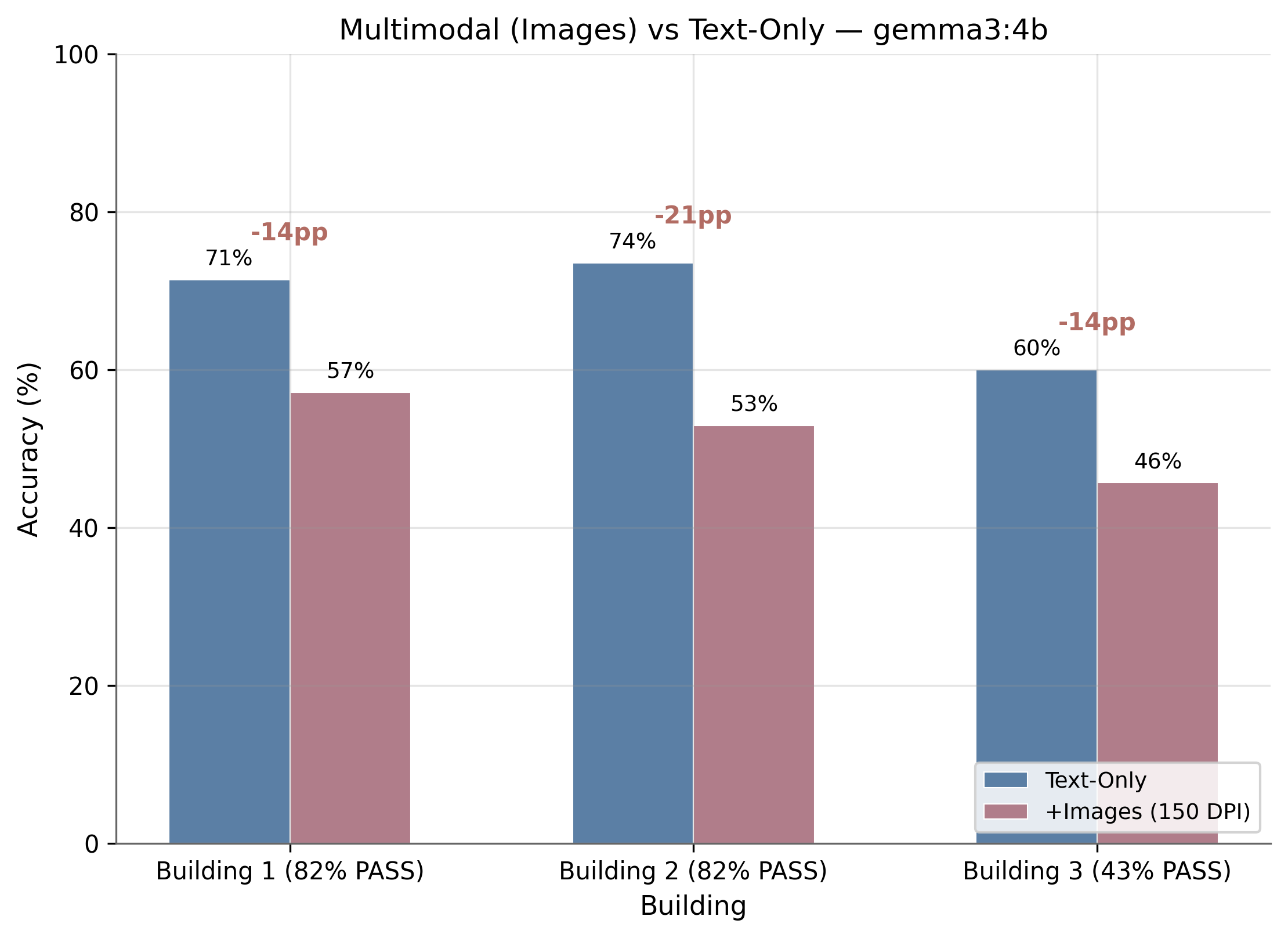}
\caption{Multimodal ablation comparing text-only gemma3:4b with +Images at 150 DPI. Adding images reduces accuracy by 14--21~pp across all three buildings. Delta annotations show the per-building change in percentage points.}
\label{fig:multimodal_ablation}
\end{figure}

Table~\ref{tab:multimodal} compares the text-only gemma3:4b configuration with a multimodal configuration that supplied up to four rendered drawing images per credit at 150 DPI. Figure~\ref{fig:multimodal_ablation} provides a direct visual comparison of the two conditions. Contrary to the expectation that images would recover evidence missing from text extraction, image inputs consistently reduced accuracy.

\begin{table}[htbp]
\centering
\small
\caption{Multimodal ablation for gemma3:4b. Images were rendered at 150 DPI and attached as additional evidence inputs.}
\label{tab:multimodal}
\begin{tabular}{lcccc}
\toprule
\textbf{Building} & \textbf{Text-Only} & \textbf{+Images (150 DPI)} & \textbf{Delta} & \textbf{Images/Credit} \\
\midrule
Building 1 & 71.4\% & 57.1\% & -14.3 pp & 1.4 \\
Building 2 & 73.5\% & 52.9\% & -20.6 pp & 0.7 \\
Building 3 & 60.0\% & 45.7\% & -14.3 pp & 0.0 \\
\midrule
Average & 68.3\% & 51.9\% & -16.4 pp & 0.7 \\
\bottomrule
\end{tabular}
\end{table}

The mean loss from adding images was 16.4 percentage points. Building 2 experienced the largest decline, from 73.5\% to 52.9\%. Building 1 and Building 3 each declined by 14.3 percentage points. The consistency of the decline across buildings indicates that the effect was systematic rather than the result of one anomalous project corpus.

The most likely explanation is resolution mismatch between technical construction drawings and the visual encoder. LEED documentation includes schedules, title blocks, equipment tables, callouts, and material notes that often use small text. At 150 DPI, a 3 mm schedule character renders at approximately 1.8 pixels in height. This resolution is below the level required for reliable visual text recognition by general-purpose vision encoders. The model therefore receives an image containing graphical structure but insufficient legible text. In this condition, it may infer plausible information from lines, grids, and nearby context rather than read the underlying evidence. Such behavior can produce both false positives and false negatives.

\begin{figure}[htbp]
\centering
\includegraphics[width=\columnwidth]{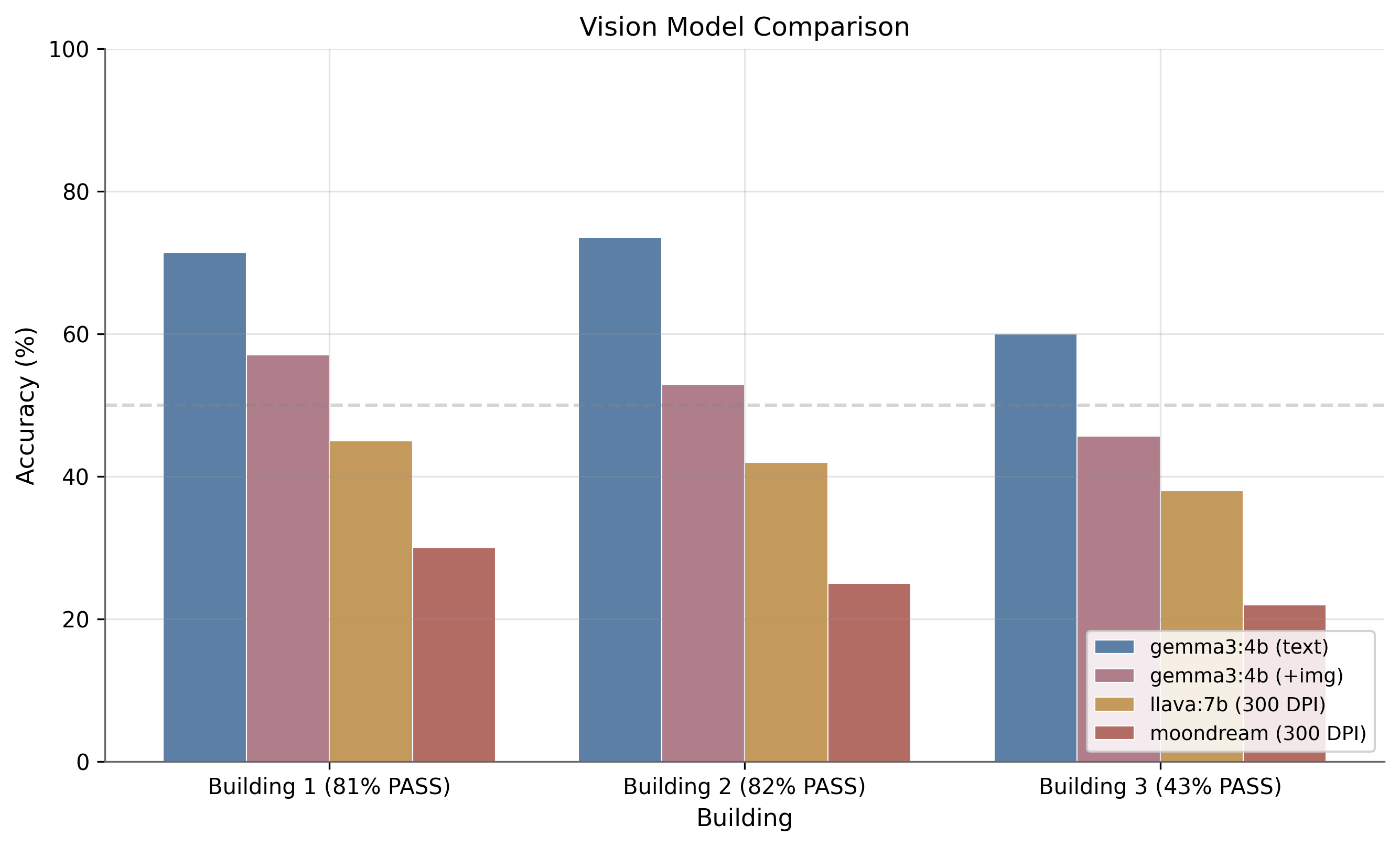}
\caption{Vision model comparison on Building~1 at 300 DPI. No vision model exceeds 50\% accuracy. text-only gemma3:4b remains the strongest configuration. llava:7b reads clean documents but fails on technical drawings; moondream hallucinates drawing content.}
\label{fig:vision_model}
\end{figure}

The effect did not reverse at 300 DPI in the additional Building 1 tests, as shown in Figure~\ref{fig:vision_model}. llava:7b could interpret clean, visually simple documents but did not reliably extract specific evidence from technical schedules or drawing sheets. moondream generated broad image descriptions, such as identifying the presence of a table or drawing, but did not reliably read numeric values or compliance-relevant text. The result suggests that simply replacing one general-purpose visual model with another does not solve the problem. The missing capability is accurate domain-specific interpretation of dense technical drawings.

OCR enrichment also did not improve accuracy. EasyOCR extracted approximately 200 to 3,800 characters per page from schedules and title blocks at 150 to 300 DPI. However, the extracted material largely duplicated text already present in the PDF text layer. Most digitally produced construction drawings embed selectable schedule text, even when that text is difficult for a vision model to read as an image. OCR therefore added noise and duplicate content rather than new evidence. The practical conclusion is clear: text-only processing is currently preferable for this corpus. Multimodal augmentation should be reconsidered only after using substantially higher-resolution rendering, targeted schedule detection, or a vision model trained specifically on construction drawings.

\subsection{Prompt Engineering Sensitivity to Class Imbalance}

\begin{figure}[htbp]
\centering
\includegraphics[width=\columnwidth]{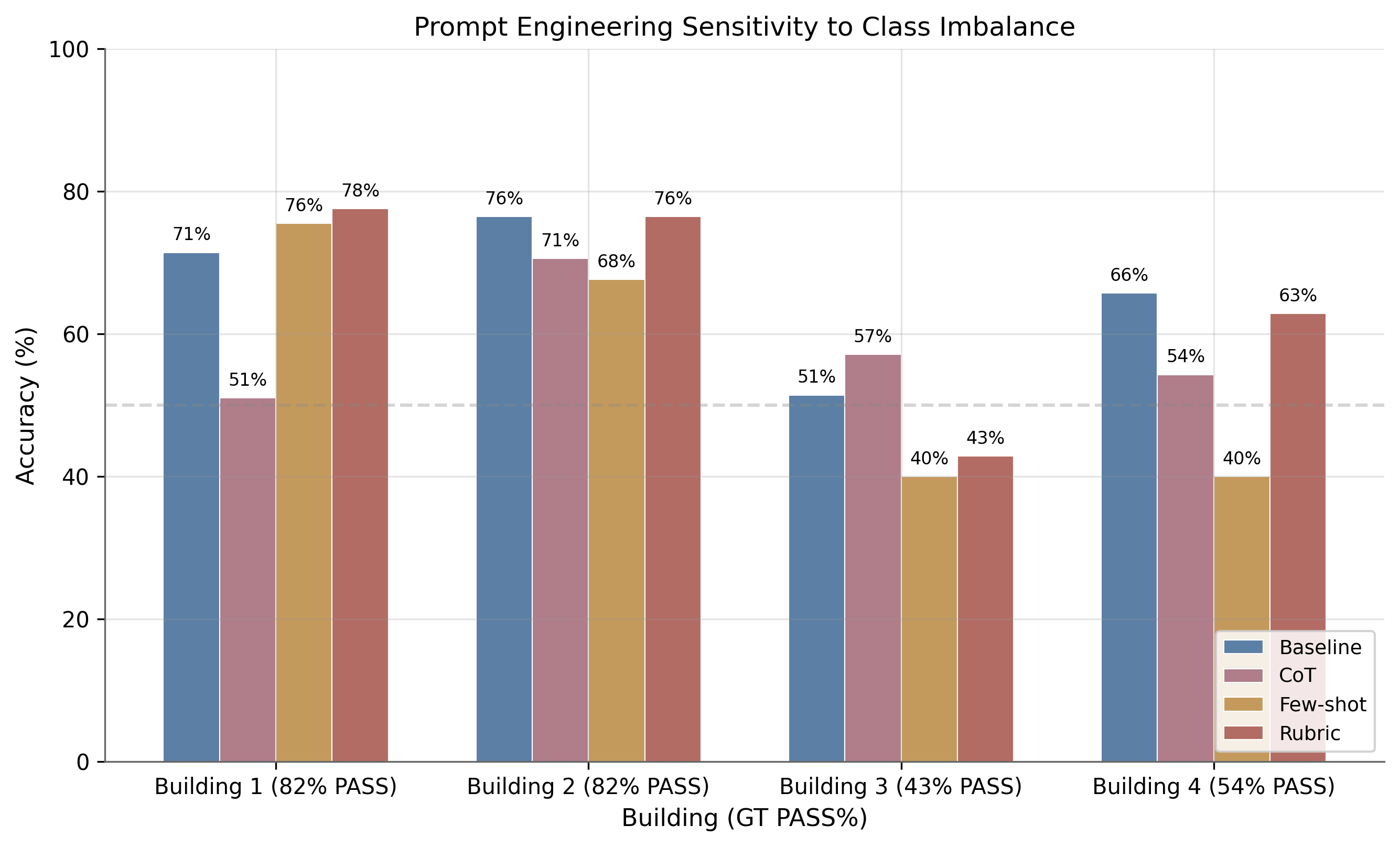}
\caption{Prompt engineering sensitivity to class imbalance. Rubric prompts perform best on PASS-heavy buildings (Building~1, 81.6\% PASS); CoT prompts perform best on INSUFFICIENT\_DATA-heavy buildings (Building~3, 42.9\% PASS). Few-shot and CoT degrade performance on balanced corpora (Building~4).}
\label{fig:prompt_engineering}
\end{figure}

\begin{figure}[htbp]
\centering
\includegraphics[width=\columnwidth]{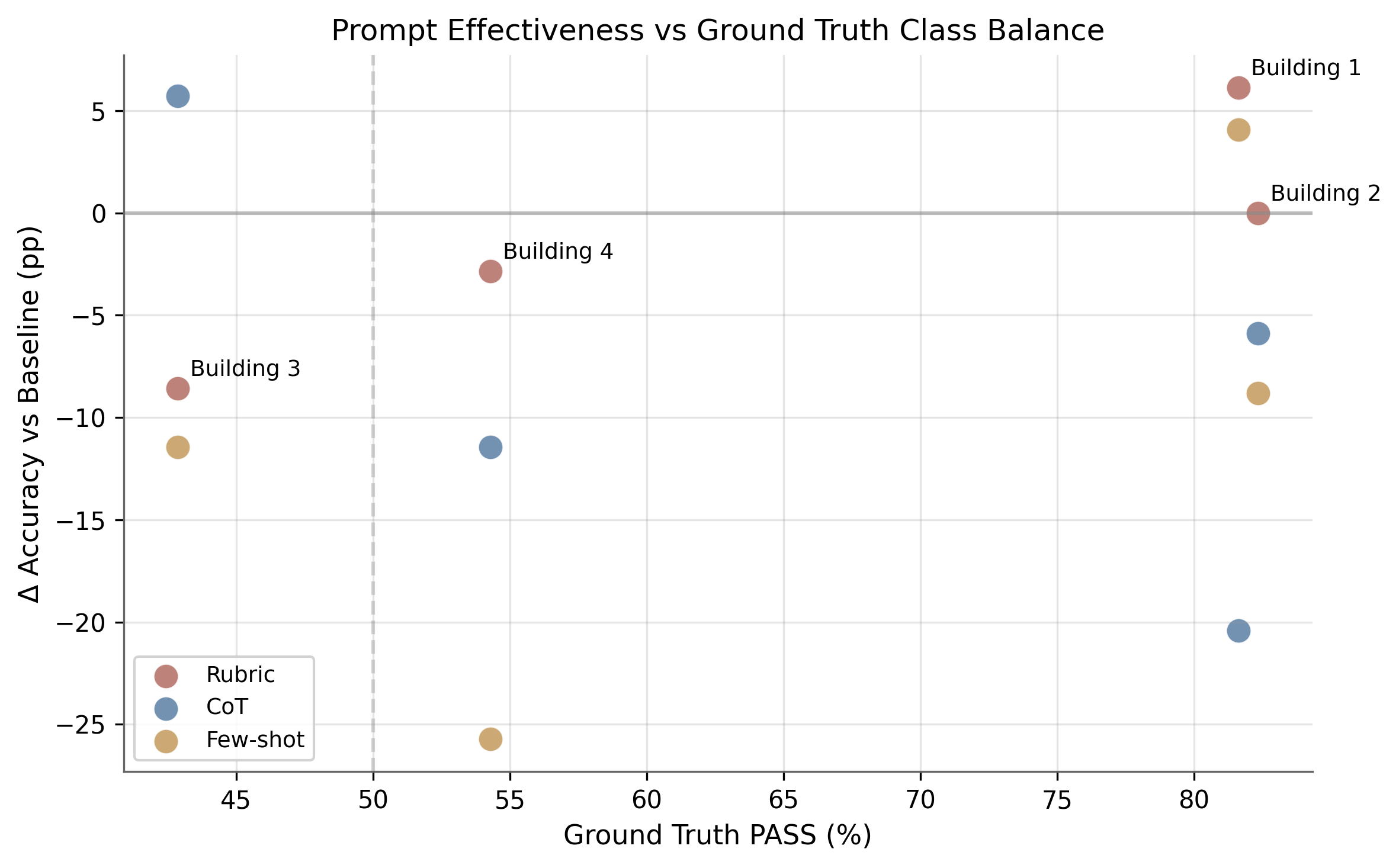}
\caption{Prompt effectiveness vs.\ ground-truth PASS rate per building. Rubric (red) shows positive accuracy deltas on PASS-heavy buildings (PASS\% $>80$); CoT (blue) shows positive deltas on INSUFFICIENT\_DATA-heavy buildings (PASS\% $<50$). The crossover occurs near 50\% PASS.}
\label{fig:class_imbalance_prompt}
\end{figure}

Table~\ref{tab:prompt_results} reports the prompt-engineering sweep for gemma3:4b using one inference run per credit. The experiment compared a baseline prompt with Chain-of-Thought (CoT), few-shot, and rubric-based prompts. The corresponding bar-chart comparison and scatter-plot analysis of class-imbalance effects are shown in Figures~\ref{fig:prompt_engineering} and~\ref{fig:class_imbalance_prompt}, respectively. Results show that no prompt strategy was universally best. Instead, prompt effectiveness depended on the distribution of PASS and \texttt{INSUFFICIENT\_DATA} labels in the building-level ground truth.

\begin{table}[htbp]
\centering
\small
\caption{Prompt-engineering results for gemma3:4b with one inference run per credit. GT PASS\% denotes the proportion of ground-truth PASS verdicts in each building.}
\label{tab:prompt_results}
\begin{tabular}{lcccccc}
\toprule
\textbf{Building} & \textbf{GT PASS\%} & \textbf{Baseline} & \textbf{CoT} & \textbf{Few-shot} & \textbf{Rubric} & \textbf{Best} \\
\midrule
Building 1 & 81.6\% & 71.4\% & 51.0\% & 75.5\% & \textbf{77.6\%} & Rubric (+6.1 pp) \\
Building 2 & 82.4\% & 76.5\% & 70.6\% & 67.6\% & 76.5\% & Baseline (+0.0 pp) \\
Building 3 & 42.9\% & 51.4\% & \textbf{57.1\%} & 40.0\% & 42.9\% & CoT (+5.7 pp) \\
Building 4 & 54.3\% & \textbf{65.7\%} & 54.3\% & 40.0\% & 62.9\% & Baseline (+0.0 pp) \\
\bottomrule
\end{tabular}
\end{table}

The rubric prompt performed best on Building 1, reaching 77.6\% accuracy. Building 1 had an 81.6\% PASS rate. In this context, explicitly presenting the criteria for sufficient evidence appears to have reduced false-negative uncertainty. The rubric prompt listed the documents, measurements, calculations, and named reports that could support a PASS decision. This structure gave the model a clearer decision boundary. Instead of treating a commissioning report or an energy-model summary as merely relevant, the model could match the evidence against an explicit criterion for sufficiency.

Building 2 had a similarly high PASS rate of 82.4\%, but the rubric prompt did not improve on the baseline. Both achieved 76.5\% accuracy. This result indicates that class distribution is influential but not sufficient by itself to predict prompt performance. Building 2's documentation may already have been clear enough that additional instruction did not improve the model's evidence interpretation. Alternatively, the rubric could have constrained the model in cases where the documentation demonstrated compliance using project-specific phrasing not fully captured by the generic criteria.

CoT performed best on Building 3, increasing accuracy from 51.4\% to 57.1\%. Building 3 had the lowest PASS rate in the dataset at 42.9\%. CoT required the model to classify evidence quality before generating the final verdict. This procedure likely increased the evidentiary threshold for PASS, which was beneficial in a corpus where many credits correctly lacked sufficient documentation. The result suggests that deliberate reasoning prompts can improve discrimination when the primary error risk is false-positive approval rather than false-negative uncertainty.

Building 4 produced the most balanced label distribution, with 54.3\% PASS. The baseline prompt was best at 65.7\%. Every prompt modification reduced accuracy. This result is consistent with the idea that each prompt introduces directional bias. Rubric prompts shift the model toward recognizing PASS evidence. CoT shifts the model toward stricter evidence requirements. Few-shot examples influence predictions toward the label distribution represented by the examples. When the underlying class distribution is relatively balanced, these directional interventions can be more harmful than a neutral baseline.

Few-shot prompting performed poorly on Building 2, Building 3, and Building 4. The examples were intentionally PASS-oriented to demonstrate evidence extraction and numeric interpretation. That design likely increased the model's effective prior probability of PASS. This was mildly useful on Building 1, where the true PASS rate was high, but harmful where the ground truth was balanced or \texttt{INSUFFICIENT\_DATA}-heavy. The finding is practically important. Few-shot prompts should not be assumed to improve compliance reasoning automatically. Their examples encode assumptions about what a typical correct outcome looks like, and those assumptions must match the deployment corpus.

Taken together, the experiment provides evidence that prompt selection should be treated as a calibration decision rather than a fixed model setting. A rubric-based prompt is appropriate when the project corpus is expected to contain complete documentation for most attempted credits. CoT is more appropriate when incomplete evidence is common. A baseline prompt may be preferable when the project has a balanced mixture of supported and unsupported credits. The results do not establish a universal causal rule because the experiment includes only four buildings, but they identify a clear and actionable interaction between prompt structure and class distribution.

\subsection{Runtime and Computational Cost}

\begin{figure}[htbp]
\centering
\includegraphics[width=\columnwidth]{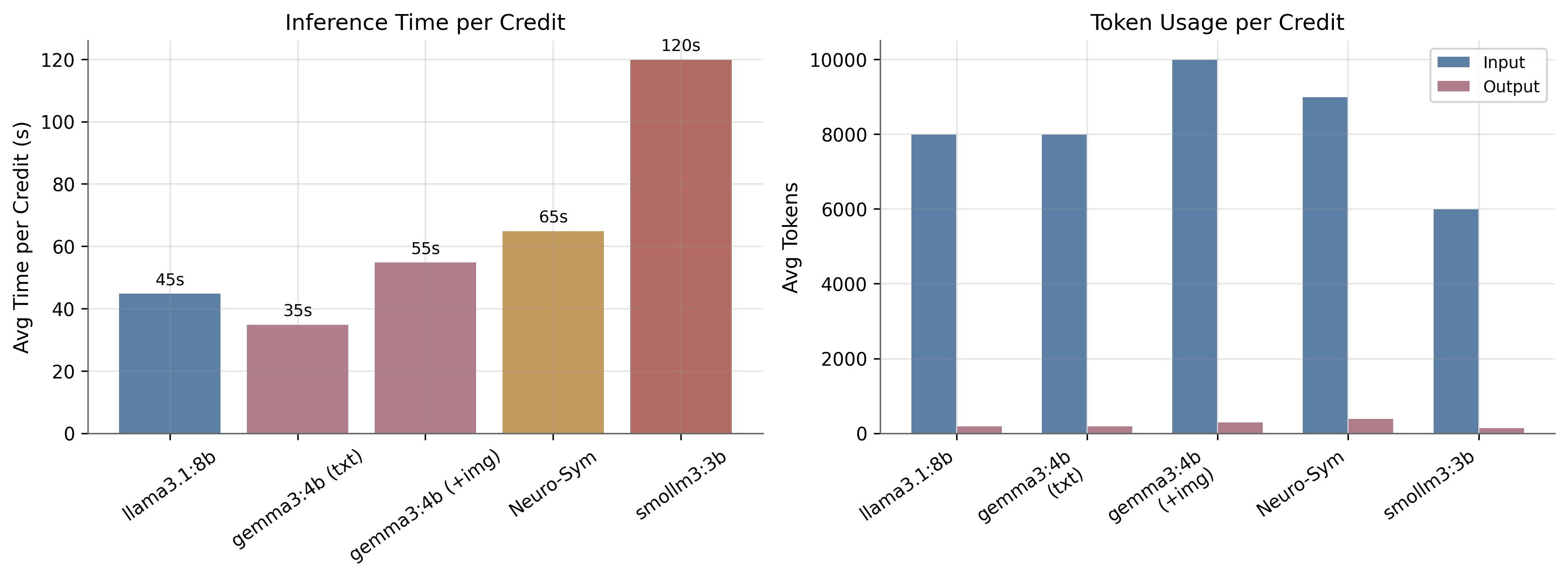}
\caption{Inference time (left) and token usage (right) per credit. Neuro-symbolic adds approximately 30~s overhead over text-only gemma3:4b. Multimodal adds 20~s without accuracy benefit. smollm3:3b is the slowest despite fewer parameters.}
\label{fig:runtime_cost}
\end{figure}

Table~\ref{tab:runtime} reports mean inference time and approximate token usage per credit, with the full comparison visualized in Figure~\ref{fig:runtime_cost}. All models were executed locally on a single NVIDIA RTX 4070 GPU with 8 GB of VRAM. No cloud model APIs were used. The local deployment requirement is significant because LEED documentation can contain proprietary drawings, contractor records, product information, and owner-sensitive building data.

\begin{table}[htbp]
\centering
\small
\caption{Runtime and approximate token cost per credit. Times include document retrieval, prompt assembly, model inference, and output parsing.}
\label{tab:runtime}
\begin{tabular}{lcccc}
\toprule
\textbf{Model} & \textbf{Time (s)} & \textbf{Input Tokens} & \textbf{Output Tokens} & \textbf{Total Tokens} \\
\midrule
llama3.1:8b (V11)      & 45  & 8,000  & 200 & 8,200 \\
gemma3:4b (V17 text)   & 35  & 8,000  & 200 & 8,200 \\
gemma3:4b (+images)    & 55  & 10,000 & 300 & 10,300 \\
Neuro-Sym (gemma3:4b)  & 65  & 9,000  & 400 & 9,400 \\
smollm3:3b             & 120 & 6,000  & 150 & 6,150 \\
\bottomrule
\end{tabular}
\end{table}

gemma3:4b text-only was the fastest effective configuration, requiring approximately 35 seconds per credit. llama3.1:8b required approximately 45 seconds, reflecting its larger parameter count. The multimodal configuration increased runtime to approximately 55 seconds per credit because image preparation, image encoding, and the larger multimodal context increased processing demand. This added cost is not justified by the observed accuracy results, since image inputs reduced performance.

The full neuro-symbolic pipeline required approximately 65 seconds per credit. The additional 30 seconds relative to text-only gemma3:4b arose from iterative evidence retrieval, structured-value extraction, deterministic numeric checking, and additional output handling. This overhead is material at the individual-credit level, but it remains modest in comparison with manual review. At 65 seconds per credit, an evaluation of 154 credit instances requires less than three hours of model runtime on a single consumer GPU, excluding document preprocessing. A manual reviewer may spend several hours locating and interpreting evidence for a single difficult credit, particularly where documentation is distributed across drawings, specifications, simulations, and contractor records.

smollm3:3b was the slowest model despite its smaller parameter count. Its lower token-processing efficiency and repeated conservative output behavior limited its practical usefulness. The result reinforces that parameter count is not a reliable proxy for either speed or effectiveness in local compliance-verification deployments.

\begin{figure}[htbp]
\centering
\includegraphics[width=\columnwidth]{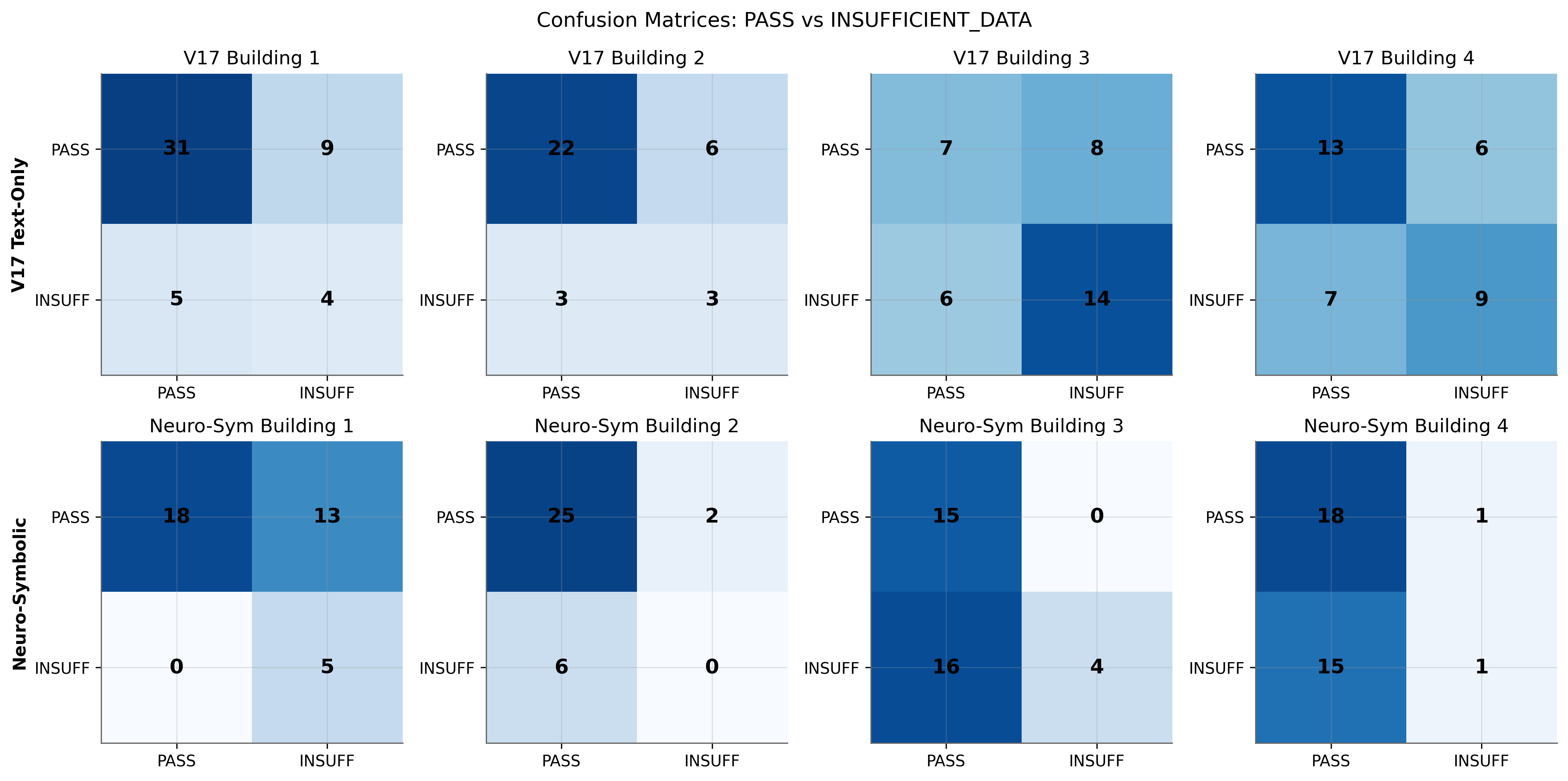}
\caption{Confusion matrices (PASS vs.\ INSUFFICIENT\_DATA) for V17 text-only (top row) and neuro-symbolic (bottom row) across all four buildings. Neuro-symbolic reduces false negatives on PASS-heavy buildings (Building~1, Building~2) via promotion logic, but introduces additional errors on Building~3 and Building~4 where ground truth is more balanced.}
\label{fig:confusion_matrices}
\end{figure}

Confusion matrices contrasting the V17 text-only and neuro-symbolic configurations are provided in Figure~\ref{fig:confusion_matrices}. These reveal that the neuro-symbolic pipeline reduces false negatives on PASS-heavy buildings (Building 1 and Building 2) via the promotion mechanism, while yielding a more mixed outcome on Building 3 and Building 4 where the ground truth is more balanced.

Overall, the runtime analysis shows that the proposed approach is operationally feasible for batch pre-review of LEED documentation. The neuro-symbolic configuration incurs an accuracy--runtime tradeoff: it is slower than text-only gemma3:4b, but it provides deterministic protection against arithmetic errors on selected quantitative credits. In practice, a hybrid deployment is likely to be most efficient: apply text-only gemma3:4b broadly, then invoke the numeric checker selectively for credits with explicit threshold logic and high certification consequence, particularly prerequisites such as EA-p2.

\section{Discussion}
\label{sec:discussion}

\subsection{Why Neuro-Symbolic Underperforms Text-Only Overall}

The main empirical result is that the neuro-symbolic pipeline, despite adding a deterministic numeric checker and confidence-based promotion, achieved lower overall accuracy (61.6\%) than text-only gemma3:4b (67.3\%). At first glance this is surprising. An exact symbolic module should improve performance, not reduce it. The explanation lies in the mismatch between where symbolic logic helps and where most credits live.

The pipeline delivers a clear gain in Energy and Atmosphere, with category accuracy increasing by 19.1 percentage points and EA-p2 rising from 50.0\% to 100.0\%. These improvements occur on credits with explicit quantitative thresholds and well-structured documentary evidence. However, only a minority of the evaluated credits have implemented numeric checkers. The majority of credits remain purely qualitative, relying on narrative evidence, complex documentation structure, and spatial or contextual interpretation. For these credits, the neuro-symbolic pipeline introduces additional steps without providing a new source of correctness.

The agentic retrieval layer is one source of degradation. Its multi-pass, metadata-filtered searches occasionally retrieve less relevant text than the simpler retrieval used in the text-only configuration, especially when credit-specific keywords are sparse or misaligned with document headings. The promotion logic is another. While it corrects conservative false negatives on PASS-heavy buildings, it can also promote genuinely uncertain cases to PASS when the underlying confidence estimate is miscalibrated. Finally, when the numeric checker is invoked with incomplete or mis-typed values, it can replace a cautious but correct language-model verdict with a confident but wrong deterministic verdict.

The conclusion is not that neuro-symbolic design is flawed. It is that the current implementation is partial. A numeric checker alone is not enough for a rating system where most credits depend on narrative, documentary, or spatial reasoning. A fuller neuro-symbolic system would need symbolic modules for qualitative requirement decomposition, document-structure parsing, and cross-credit evidence linking. In the absence of these modules, adding a symbolic layer for a small subset of credits can change retrieval and decision dynamics for the whole pipeline in ways that reduce aggregate performance.

The practical implication is to deploy a hybrid rather than the full pipeline. Text-only gemma3:4b should remain the default verifier for all credits because it is both faster (about 35 seconds per credit) and more robust on qualitative categories. The numeric checker should be applied as a post-processing overlay on the subset of credits where deterministic rules have been implemented and extraction reliability has been validated. This design keeps symbolic rigor where it clearly helps and avoids introducing complexity where it does not.

\subsection{Numeric Checker: Successes, Limitations, and Extraction Dependency}

The numeric checker behaves exactly as intended when it receives valid inputs. On credits where the required quantities are correctly extracted, its application of LEED threshold logic is exact, and it does not introduce new errors. EA-p2 illustrates this clearly. Once the energy-improvement percentage was reliably recovered from energy-model documentation, the checker moved accuracy from 50.0\% to 100.0\%. The same pattern holds for EA-c5, MR-c5, EQ-p1, and LT-c5, where clean values and straightforward rules produced 25.0 percentage-point gains.

The main limitation is not the rules themselves. It is reliable extraction of semantically correct, correctly typed values from heterogeneous documents. In the current experiments, a substantial share of checker invocations returned \texttt{INSUFFICIENT\_DATA} because the required values were not available in the extracted text in a usable form. In those cases, the checker defers back to the language model. The net impact of the checker on accuracy is therefore driven by the coverage and quality of the extraction step, not by the correctness of the threshold logic, which is exact for the credits it supports.

Three extraction failure modes were common. The first is simple absence: the quantity the checker expects, such as a water-use reduction percentage or daylight autonomy metric, does not appear as a single explicit statement in the retrieved text. It may be implicit in a table, spread across multiple documents, or presented only as raw inputs rather than as a summarized result. The second is type confusion: a numeric value is extracted, but it corresponds to a different concept than the checker expects. WE-c2 showed this behavior when airflow rates were mistaken for water-reduction percentages. The third is context mismatch: the value is correct but the surrounding assumptions are missing, so the checker cannot reliably decide whether the threshold applies. For EQ-c7, a reported sDA value might be valid for a simulation zone that is not the occupied area relevant to the credit.

These are engineering problems rather than structural limits of neuro-symbolic design. They point to three directions for improvement. First, credit-specific extraction templates can define not only what to extract but also the expected units, typical ranges, source-document types, and sanity checks. Second, table-aware parsing can treat schedules and tabular data as structured objects instead of linear text, extracting key-value pairs directly from table cells. Many of the quantities needed for LEED verification live in tables rather than in narratives. Third, cross-document reconciliation can detect and compare repeated values across documents, flagging discrepancies rather than choosing one arbitrarily.

The dependency on extraction also limits how far the numeric checker can be extended. Each additional credit requires understanding where its evidence lives and how it is expressed. Extending to more credits will be worthwhile only if extraction success rates are high enough to justify building and maintaining new rule modules.

\subsection{Multimodal Processing: Resolution as a Binding Constraint}

The multimodal experiments showed a consistent and large negative effect. Adding rendered drawing images at 150 dpi reduced accuracy by 14 to 21 percentage points across all evaluated buildings. This runs directly against the intuitive expectation that visual evidence should complement text. The most plausible explanation is that the pipeline delivers visual inputs at a resolution that is too low for reliable reading of technical content.

Construction drawings carry critical evidence in small text: equipment schedules, title blocks, note annotations, and detail labels. At 150 dpi, typical schedule text renders at about two pixels of character height. General-purpose vision encoders are not designed to read text at that resolution. They perceive lines, boxes, and overall layout but not the specific characters. In practice, this means the model sees structured images but cannot see the numbers and words that matter. When asked to reason about credit compliance, it produces descriptions and inferences that are loosely grounded in the visual impression rather than in actual text, creating both false positives and false negatives.

Increasing resolution to 300 dpi does not fully solve the problem. Character height increases, but not enough to reach the 10-pixel scale where most vision models begin to read text reliably. Larger titles and sheet labels become more legible, yet these seldom carry the detailed numeric evidence required for threshold-based decisions. The net effect in the current tests is still negative.

OCR enrichment behaves similarly. While OCR can extract text from drawings, much of that text already exists in the PDF’s own text layer for digitally produced schedules. OCR tends to add duplicates and noise rather than genuinely new evidence. In this pipeline, it did not translate into measurable accuracy gains.

For now, the practical conclusion is straightforward. Text-only processing is the safer default for LEED documentation that includes technical drawings. Multimodal augmentation should be considered only if three conditions change: a vision model is trained specifically on construction drawings, rendering resolution is increased substantially, or the pipeline incorporates a targeted module that isolates and upscales evidence-heavy regions such as schedule tables before passing them to the visual encoder. Until then, adding images at standard document resolutions is more likely to harm than help.

\subsection{Prompt Sensitivity: Bayesian Interpretation and Deployment Implications}

Prompt experiments show that prompt choice is not neutral. It interacts with the underlying distribution of PASS and \texttt{INSUFFICIENT\_DATA} labels in the building under review. This interaction can be understood as a prompt-induced shift in the model’s effective prior.

On PASS-heavy buildings, rubric prompts performed best. GLT and HMA both had PASS rates above 80\%. On GLT, a rubric prompt that lists explicit PASS criteria raised accuracy from 71.4\% to 77.6\%. The rubric acts like a structured prior that says: if you see these forms of evidence, treat them as strong indications of compliance. In a context where most credits genuinely have adequate documentation, this shift reduces unnecessary hesitation and corrects conservative false negatives.

On buildings with lower PASS rates, Chain-of-Thought prompts did better. RRH, with about 43\% PASS, showed a gain from 51.4\% to 57.1\% under CoT. CoT forces the model to categorize evidence and to justify its verdict step by step. This effectively raises the threshold for calling a credit PASS. In a documentation-lean setting, this more cautious stance reduces false-positive approvals.

Few-shot prompts behaved as their examples suggest. Because the examples were mostly PASS cases, they raised the model’s effective prior probability of PASS. This was tolerable on GLT but harmful elsewhere. When the true PASS rate diverged substantially from the example distribution, the model was nudged toward over-approval.

The baseline prompt performed best on SEA, where the PASS rate was closer to 50\%. In such a balanced regime, prompts that push the model toward either more PASS or more \texttt{INSUFFICIENT\_DATA} can skew performance away from the ground truth.

Taken together, these results support a simple guideline. Before selecting a prompt, project teams should have an estimate of how many credits are genuinely well documented. If most attempted credits are expected to pass, a rubric-style prompt is appropriate. If many credits are realistically borderline or unsupported, a CoT-style prompt is safer. If the expected distribution is roughly balanced, sticking with a neutral baseline prompt is more likely to avoid systematic bias. This guideline does not replace project-specific validation, but it provides a principled starting point.

\subsection{Limitations}

Several limitations constrain how these findings should be interpreted.

First, the corpus is small. The analysis covers 153 credit-level decisions across four buildings. This is enough to see patterns and to generate hypotheses, but not enough to make strong statistical claims about how the system will behave in all LEED projects. Accuracy figures are descriptive for this dataset. They are not precise estimates of broader performance.

Second, there are no verified FAIL labels. Ground truth consists only of PASS and \texttt{INSUFFICIENT\_DATA}. This reflects common practice: project teams rarely submit documentation for credits they know they will fail. As a result, the study cannot measure how well the system detects genuine non-compliance. Any incorrect PASS on a truly non-compliant credit would go unobserved in this dataset.

Third, prompt comparisons were conducted with single runs per credit, without self-consistency voting. This was necessary to keep computation tractable, but it means the observed prompt differences include both real shifts and sampling noise. Running multiple samples per prompt and credit could refine the comparison.

Fourth, the numeric checker is intentionally narrow. It covers 12 quantitative credits as they appear in the evaluated buildings. Extending it to more credits, more rating-system variants, or different project types will require careful rule implementation and version control. The current work demonstrates feasibility, not completeness.

Fifth, the promotion threshold was chosen as a single global value. It was not tuned separately per building using held-out data. The reported gains therefore represent an optimistic bound rather than a fully validated deployment setting. A more rigorous calibration would estimate thresholds using development credits and then test them on previously unseen credits within the same building.

Finally, the pipeline assumes that PDF text extraction is reliable. In the evaluated corpus, most documents were digitally produced and had intact text layers. Scanned documents, image-only PDFs, or poorly extracted files would change this assumption. The study does not yet evaluate performance under those conditions.

\subsection{Implications for Practice}

Even with these limits, several practical lessons emerge.

The first is model choice. For now, text-only gemma3:4b is the most dependable core verifier. It balances accuracy, speed, and local deployability. Smaller models that default heavily to \texttt{INSUFFICIENT\_DATA} are not acceptable for documentation-rich projects.

The second is how to use symbolic checking. The numeric checker should be treated as a targeted tool, not a universal layer. Its value is clearest on EA-p2 and a small set of other quantitative credits where arithmetic errors are costly and evidence is well structured. Applying it as a post-processing step, rather than coupling it tightly to all verdicts, reduces the risk that its failures propagate.

The third is prompt selection. Prompts behave like tunable priors. Choosing them with some awareness of expected documentation strength—PASS-heavy, balanced, or documentation-lean—helps align model behavior with project reality.

The fourth is modality. For the kinds of drawings and schedules seen here, adding images at common resolutions reduces accuracy. It should be avoided in production until the vision stack is demonstrably ready for dense technical drawings.

The fifth is where to invest engineering effort. The main bottleneck now is extraction: finding and structuring the right numbers and statements from complex document sets. Improvements in extraction will feed both the language model and the symbolic checker. They are likely to yield more benefit than further prompt tuning or incremental changes in model size.

\subsection{Future Work}

Several next steps follow directly from the current findings.

A larger, more diverse benchmark is a priority. A corpus that covers more projects, more building types, more rating-system versions, and a realistic mix of PASS, FAIL, and \texttt{INSUFFICIENT\_DATA} credits would turn these descriptive results into stronger evidence. It would also enable cross-group comparison and reduce evaluation fragmentation.

Expanding symbolic coverage beyond numeric thresholds is another. Many LEED credits can be decomposed into checkable predicates about documents and processes, even when they do not boil down to a single number. Symbolic modules that encode those predicates and the relationships between them would extend the benefits of neuro-symbolic design to a larger share of the rating system.

Improving document-structure awareness is a third direction. Treating schedules, tables, and forms as structured objects rather than linear text would improve extraction of the very values the numeric checker needs most.

Building cross-credit evidence graphs is a fourth. Evidence for one credit often supports others. Linking documents and extracted values to all relevant credits would reduce missed evidence and lower the rate of \texttt{INSUFFICIENT\_DATA} verdicts caused by narrow retrieval.

Calibrating confidence and promotion per building rather than globally is a fifth. Thresholds tuned to building-specific development data are more likely to perform reliably across diverse projects than a single universal setting.

Finally, integrating with human review workflows is essential. Accuracy matters, but so do time savings, reviewer confidence, and the clarity of evidence trails. Controlled studies with LEED professionals can evaluate whether the system actually helps them work faster, more consistently, and with fewer oversights.

\section{Conclusion}
\label{sec:conclusion}

This paper introduced a neuro-symbolic pipeline for LEED v4.1 BD+C compliance checking over real project documentation. The system combines credit-aware chunking, structured retrieval, self-consistent language-model inference, deterministic numeric threshold evaluation, and calibrated promotion of conservative uncertainty.

On four UT Austin buildings, comprising hundreds of PDFs and more than a hundred credit-level decisions, the experiments show several things. A 4-billion-parameter model, gemma3:4b, is strong enough to serve as a local core verifier and outperforms a larger 8-billion-parameter model in this specific task. A deterministic numeric checker can fully correct arithmetic errors on a key prerequisite credit, EA-p2, moving it from half correct to entirely correct. Confidence-based promotion can offset the model’s tendency to overuse \texttt{INSUFFICIENT\_DATA}, especially on projects where most credits truly have adequate documentation. Adding images at typical resolutions harms performance rather than helping it, highlighting that current vision-language tools are not yet ready for dense technical drawings in this context. Prompt choice matters and interacts with the distribution of PASS and \texttt{INSUFFICIENT\_DATA} labels, suggesting that prompts should be selected with knowledge of the project’s documentation profile.

The neuro-symbolic configuration does not yet beat the best text-only baseline on aggregate accuracy, but it does solve the hardest part of the problem: making quantitative threshold decisions deterministic instead of leaving them to probabilistic text generation. The work points clearly to where effort should go next: richer symbolic treatment of qualitative requirements, better document-structure parsing, more robust extraction, and calibration that respects differences between buildings.

For practitioners, the near-term advice is simple. Use a strong, locally deployable text model as the default. Add deterministic checking carefully where arithmetic matters most. Choose prompts with awareness of how complete the documentation is likely to be. Hold off on multimodal inputs until the vision stack catches up. And treat extraction as the main engineering frontier. The combination of these steps can make LEED pre-review faster and more consistent without compromising the standards that certification is meant to uphold.


\bibliographystyle{cas-model2-names}

\bibliography{references}

@article{kalai_why_2025,
	title = {Why {Language} {Models} {Hallucinate}},
	language = {en},
	publisher = {OpenAI},
	author = {Kalai, Adam Tauman and Nachum, Ofir and Vempala, Santosh S and Zhang, Edwin},
	month = sep,
	year = {2025},
}

@inproceedings{geng_survey_2024,
	address = {Mexico City, Mexico},
	title = {A {Survey} of {Confidence} {Estimation} and {Calibration} in {Large} {Language} {Models}},
	url = {https://aclanthology.org/2024.naacl-long.366/},
	doi = {10.18653/v1/2024.naacl-long.366},
	urldate = {2026-07-14},
	booktitle = {Proceedings of the 2024 {Conference} of the {North} {American} {Chapter} of the {Association} for {Computational} {Linguistics}: {Human} {Language} {Technologies} ({Volume} 1: {Long} {Papers})},
	publisher = {Association for Computational Linguistics},
	author = {Geng, Jiahui and Cai, Fengyu and Wang, Yuxia and Koeppl, Heinz and Nakov, Preslav and Gurevych, Iryna},
	editor = {Duh, Kevin and Gomez, Helena and Bethard, Steven},
	month = jun,
	year = {2024},
	pages = {6577--6595},
}

@misc{michael_confidence_2026,
	title = {Confidence {Calibration} in {Large} {Language} {Models}},
	url = {http://arxiv.org/abs/2605.23909},
	doi = {10.48550/arXiv.2605.23909},
	urldate = {2026-07-14},
	publisher = {arXiv},
	author = {Michael, Noam and BenShushan, Daniel and Bien, Jacob and Moore, Don A.},
	month = apr,
	year = {2026},
	note = {arXiv:2605.23909 [cs.AI]
version: 1},
}

@inproceedings{huang_mirror-consistency_2024,
	address = {Miami, Florida, USA},
	title = {Mirror-{Consistency}: {Harnessing} {Inconsistency} in {Majority} {Voting}},
	shorttitle = {Mirror-{Consistency}},
	url = {https://aclanthology.org/2024.findings-emnlp.135/},
	doi = {10.18653/v1/2024.findings-emnlp.135},
	urldate = {2026-07-14},
	booktitle = {Findings of the {Association} for {Computational} {Linguistics}: {EMNLP} 2024},
	publisher = {Association for Computational Linguistics},
	author = {Huang, Siyuan and Ma, Zhiyuan and Du, Jintao and Meng, Changhua and Wang, Weiqiang and Lin, Zhouhan},
	editor = {Al-Onaizan, Yaser and Bansal, Mohit and Chen, Yun-Nung},
	month = nov,
	year = {2024},
	pages = {2408--2420},
}

@inproceedings{taubenfeld_confidence_2025,
	title = {Confidence {Improves} {Self}-{Consistency} in {LLMs}},
	url = {http://arxiv.org/abs/2502.06233},
	doi = {10.18653/v1/2025.findings-acl.1030},
	urldate = {2026-07-14},
	booktitle = {Findings of the {Association} for {Computational} {Linguistics}: {ACL} 2025},
	author = {Taubenfeld, Amir and Sheffer, Tom and Ofek, Eran and Feder, Amir and Goldstein, Ariel and Gekhman, Zorik and Yona, Gal},
	year = {2025},
	note = {arXiv:2502.06233 [cs.CL]},
	pages = {20090--20111},
}

@misc{jadon_enhancing_2025,
	title = {Enhancing {Domain}-{Specific} {Retrieval}-{Augmented} {Generation}: {Synthetic} {Data} {Generation} and {Evaluation} using {Reasoning} {Models}},
	shorttitle = {Enhancing {Domain}-{Specific} {Retrieval}-{Augmented} {Generation}},
	url = {http://arxiv.org/abs/2502.15854},
	doi = {10.48550/arXiv.2502.15854},
	urldate = {2026-07-14},
	publisher = {arXiv},
	author = {Jadon, Aryan and Patil, Avinash and Kumar, Shashank},
	month = feb,
	year = {2025},
	note = {arXiv:2502.15854 [cs.LG]},
}

@misc{oche_systematic_2025,
	title = {A {Systematic} {Review} of {Key} {Retrieval}-{Augmented} {Generation} ({RAG}) {Systems}: {Progress}, {Gaps}, and {Future} {Directions}},
	shorttitle = {A {Systematic} {Review} of {Key} {Retrieval}-{Augmented} {Generation} ({RAG}) {Systems}},
	url = {http://arxiv.org/abs/2507.18910},
	doi = {10.48550/arXiv.2507.18910},
	urldate = {2026-07-14},
	publisher = {arXiv},
	author = {Oche, Agada Joseph and Folashade, Ademola Glory and Ghosal, Tirthankar and Biswas, Arpan},
	month = jul,
	year = {2025},
	note = {arXiv:2507.18910 [cs.CL]
version: 1},
}

@article{wan_exploring_2025,
	title = {Exploring {Gen}-{AI} applications in building research and industry: {A} review},
	volume = {18},
	issn = {1996-3599, 1996-8744},
	shorttitle = {Exploring {Gen}-{AI} applications in building research and industry},
	url = {http://arxiv.org/abs/2410.01098},
	doi = {10.1007/s12273-025-1279-x},
	number = {6},
	urldate = {2026-07-14},
	journal = {Building Simulation},
	author = {Wan, Hanlong and Zhang, Jian and Chen, Yan and Xu, Weili and Feng, Fan},
	month = jun,
	year = {2025},
	note = {arXiv:2410.01098 [cs.AI]},
	pages = {1251--1273},
}

@misc{madireddy_large_2025,
	title = {Large {Language} {Model}-{Driven} {Code} {Compliance} {Checking} in {Building} {Information} {Modeling}},
	url = {http://arxiv.org/abs/2506.20551},
	doi = {10.48550/arXiv.2506.20551},
	urldate = {2026-07-14},
	publisher = {arXiv},
	author = {Madireddy, Soumya and Gao, Lu and Din, Zia and Kim, Kinam and Senouci, Ahmed and Han, Zhe and Zhang, Yunpeng},
	month = jun,
	year = {2025},
	note = {arXiv:2506.20551 [cs.SE]},
}

@article{hettiarachchi_code-accord_2025,
	title = {{CODE}-{ACCORD}: {A} {Corpus} of building regulatory data for rule generation towards automatic compliance checking},
	volume = {12},
	copyright = {2025 The Author(s)},
	issn = {2052-4463},
	shorttitle = {{CODE}-{ACCORD}},
	url = {https://www.nature.com/articles/s41597-024-04320-x},
	doi = {10.1038/s41597-024-04320-x},
	language = {en},
	number = {1},
	urldate = {2026-07-14},
	journal = {Scientific Data},
	publisher = {Nature Publishing Group},
	author = {Hettiarachchi, Hansi and Dridi, Amna and Gaber, Mohamed Medhat and Parsafard, Pouyan and Bocaneala, Nicoleta and Breitenfelder, Katja and Costa, Gonçal and Hedblom, Maria and Juganaru-Mathieu, Mihaela and Mecharnia, Thamer and Park, Sumee and Tan, He and Tawil, Abdel-Rahman H. and Vakaj, Edlira},
	month = jan,
	year = {2025},
	pages = {170},
}

@article{zhang_semantic_2016,
	title = {Semantic {NLP}-{Based} {Information} {Extraction} from {Construction} {Regulatory} {Documents} for {Automated} {Compliance} {Checking}},
	volume = {30},
	url = {https://ascelibrary.org/doi/10.1061/%28ASCE%29CP.1943-5487.0000346},
	doi = {10.1061/(ASCE)CP.1943-5487.0000346},
	language = {en},
	number = {2},
	urldate = {2026-07-14},
	journal = {Journal of Computing in Civil Engineering},
	publisher = {American Society of Civil Engineers},
	author = {Zhang, Jiansong and El-Gohary, Nora M.},
	month = mar,
	year = {2016},
	pages = {04015014},
}

@article{lee_automated_2026,
	title = {Automated compliance checking across the building lifecycle: {Systematic} and semantic review integrating {PRISMA} and deep search},
	volume = {185},
	issn = {0926-5805},
	shorttitle = {Automated compliance checking across the building lifecycle},
	url = {https://www.sciencedirect.com/science/article/pii/S0926580526001007},
	doi = {10.1016/j.autcon.2026.106859},
	urldate = {2026-07-14},
	journal = {Automation in Construction},
	author = {Lee, Ju Hyun and Ji, Seung Yeul and Ostwald, Michael J.},
	month = may,
	year = {2026},
	pages = {106859},
}

@article{mirhosseini_systematic_2026,
	title = {A systematic review of methods for interpreting building code regulations in automated compliance systems},
	volume = {0},
	issn = {0961-3218},
	url = {https://doi.org/10.1080/09613218.2026.2637965},
	doi = {10.1080/09613218.2026.2637965},
	number = {0},
	urldate = {2026-07-14},
	journal = {Building Research \& Information},
	publisher = {Routledge},
	author = {Mirhosseini, Nikoo and Shojaei, Davood and Sabri, Soheil},
	month = mar,
	year = {2026},
	note = {\_eprint: https://doi.org/10.1080/09613218.2026.2637965},
	pages = {1--24},
}

@article{feijao_comparative_2024,
	title = {Comparative analysis of sustainable building certification processes},
	volume = {96},
	issn = {2352-7102},
	url = {https://www.sciencedirect.com/science/article/pii/S2352710224019697},
	doi = {10.1016/j.jobe.2024.110401},
	urldate = {2026-07-09},
	journal = {Journal of Building Engineering},
	author = {Feijão, David and Reis, Cristina and Marques, Margarida Correia},
	month = nov,
	year = {2024},
	pages = {110401},
}

@misc{lee_integrated_2025,
	title = {An {Integrated} {Platform} for {LEED} {Certification} {Automation} {Using} {Computer} {Vision} and {LLM}-{RAG}},
	url = {http://arxiv.org/abs/2506.00888},
	doi = {10.48550/arXiv.2506.00888},
	urldate = {2026-07-09},
	publisher = {arXiv},
	author = {Lee, Jooyeol},
	month = jun,
	year = {2025},
	note = {arXiv:2506.00888 [cs.SE]},
}



\end{document}